\documentclass[twocolumn]{article}
\usepackage{preprint}
\usepackage[T1]{fontenc}
\usepackage[utf8]{inputenc}

\usepackage{amsmath,amssymb}
\usepackage{graphicx,booktabs,array,tabularx}
\usepackage[numbers,sort&compress]{natbib}
\usepackage{microtype}
\usepackage{xcolor}
\definecolor{linkink}{HTML}{315D78}
\usepackage[colorlinks=true,allcolors=linkink]{hyperref}
\usepackage{url}
\newcolumntype{L}[1]{>{\raggedright\arraybackslash}p{#1}}
\newcolumntype{Y}{>{\raggedright\arraybackslash}X}
\newcommand{\system}{Labference}
\newcommand{\legal}{\mathcal{U}}
\newcommand{\ind}{\mathbf{1}}
\newcommand{\code}[1]{\texttt{\small #1}}
\hypersetup{
 pdftitle={Labference: Benchmarking the Reasoning\textendash Action Connection in Robotic Laboratories},
 pdfauthor={},
 pdfsubject={Technical Report: state-dependent operation evaluation for VLA models and agents},
 pdfkeywords={robot manipulation, laboratory automation, reasoning, vision-language-action models, agents, online evaluation, oracle diagnostics}
}
\title{Labference: Benchmarking the Reasoning--Action\\Connection in Robotic Laboratories}
\author{}
\makeatletter
\renewcommand{\@maketitle}{%
 \vbox{\hsize\textwidth\linewidth\hsize\centering
 {\LARGE\bfseries\@title\par}
 \vskip 10pt
 {\small\textsc{Technical Report}\quad November 2, 2026\par}
 \vskip 5pt
 {\small Project page: \url{https://github.com/LabBuddy26/Labference}\par}
 \vskip 14pt}}
\makeatother
\begin{document}
\twocolumn[\begin{@twocolumnfalse}
\maketitle
\begingroup
\centering
{\small\textsc{Main Contributors}\par}
\vspace{4pt}
\newcommand{\contributor}[2]{%
  \begin{minipage}[t]{0.30\textwidth}
    \centering\vspace{0pt}
    \parbox[c][12mm][c]{\linewidth}{\centering
      \includegraphics[width=0.90\linewidth,height=10mm,keepaspectratio,#2]{media/contributors/#1}}
  \end{minipage}}
\noindent
\contributor{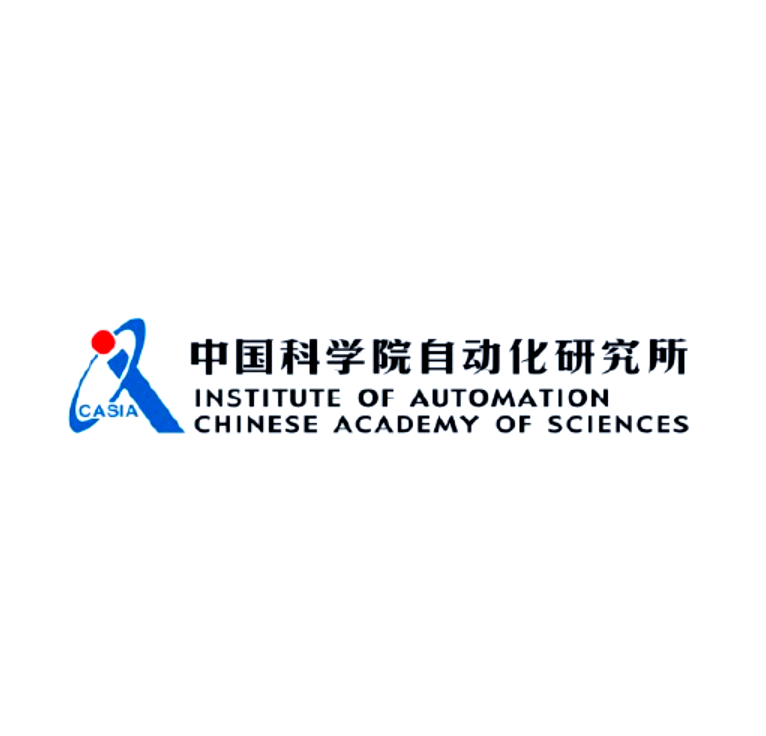}{trim=55bp 285bp 55bp 305bp,clip}\hfill
\contributor{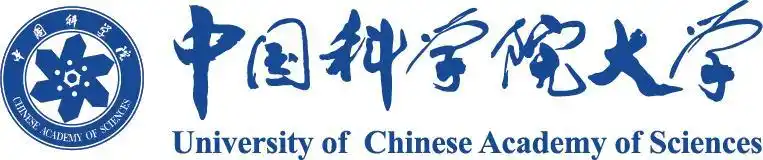}{}\hfill
\contributor{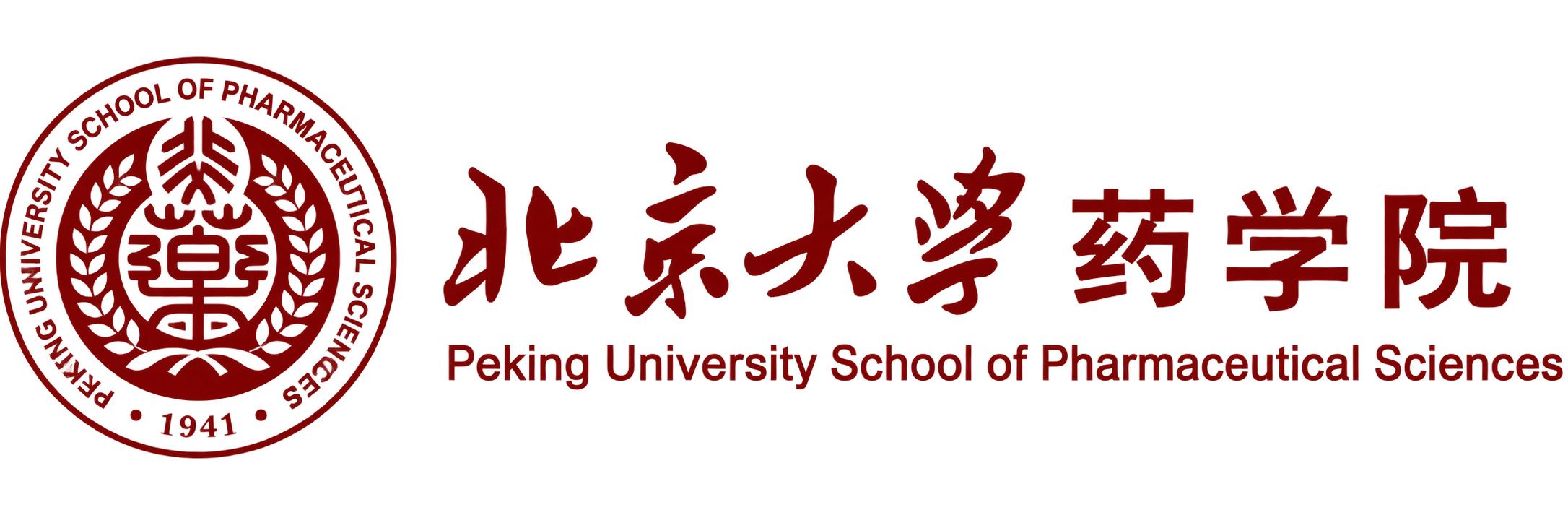}{}
\par\vspace{5pt}
\noindent
\contributor{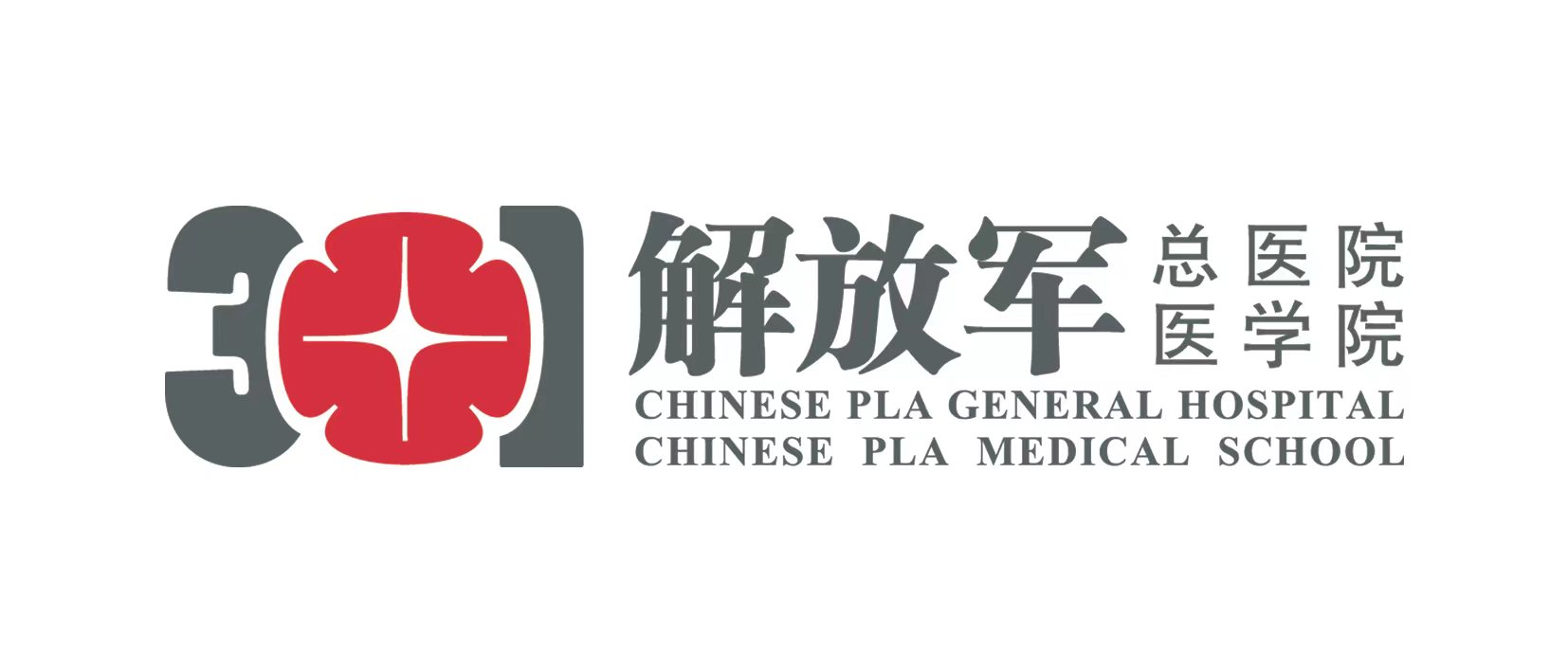}{trim=160bp 200bp 150bp 200bp,clip}\hfill
\contributor{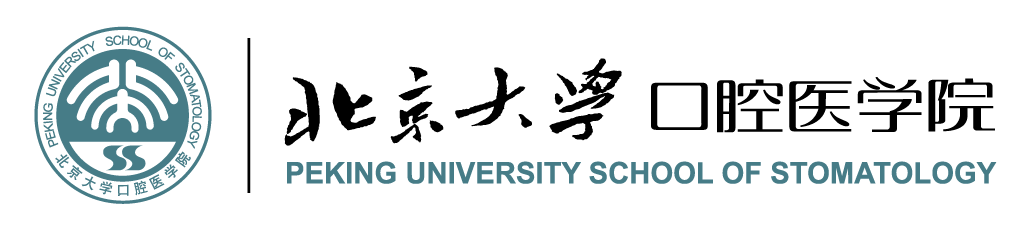}{}\hfill
\contributor{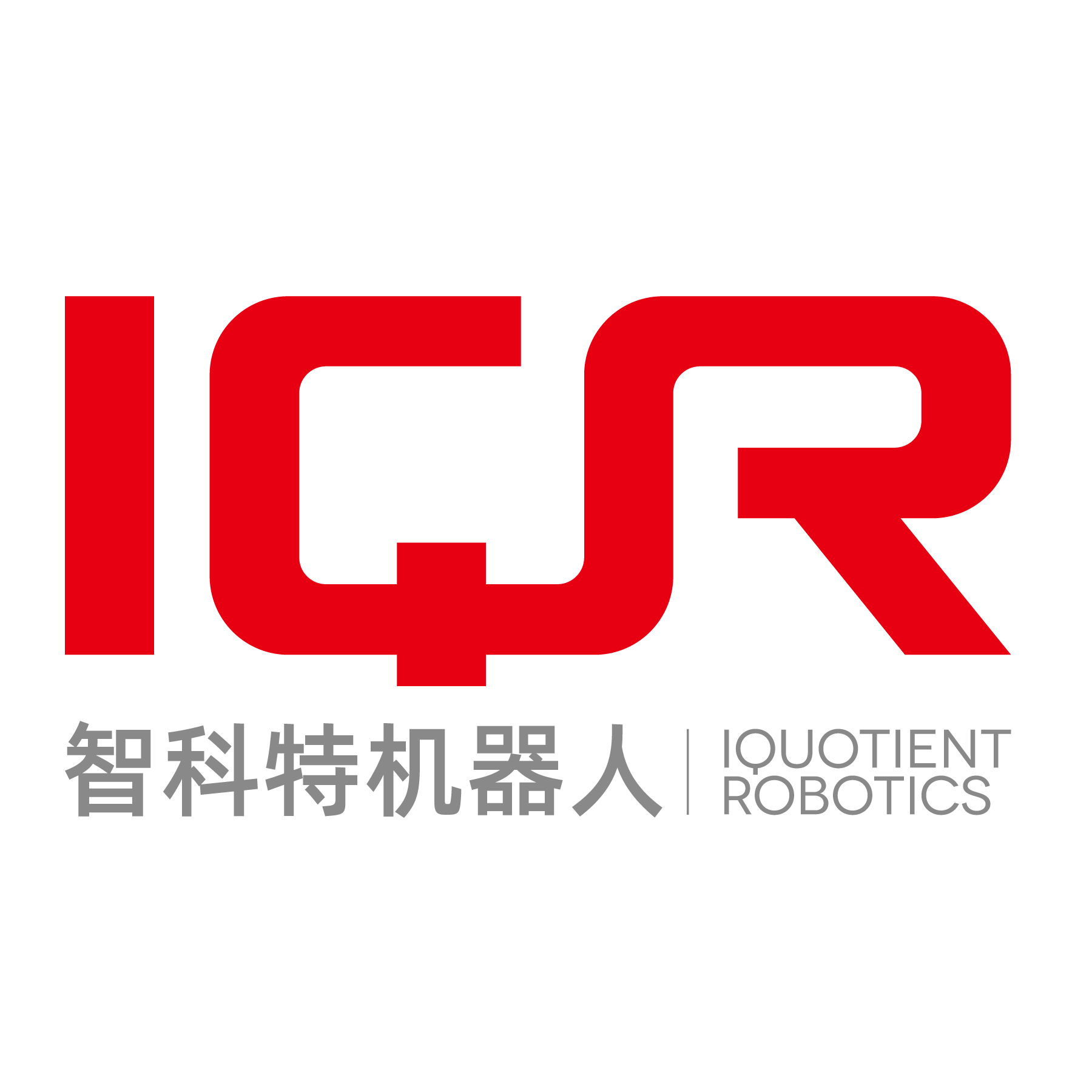}{trim=45bp 240bp 45bp 270bp,clip}
\par\vspace{5pt}
\noindent
\contributor{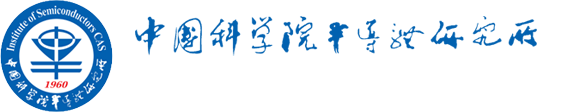}{}\hfill
\contributor{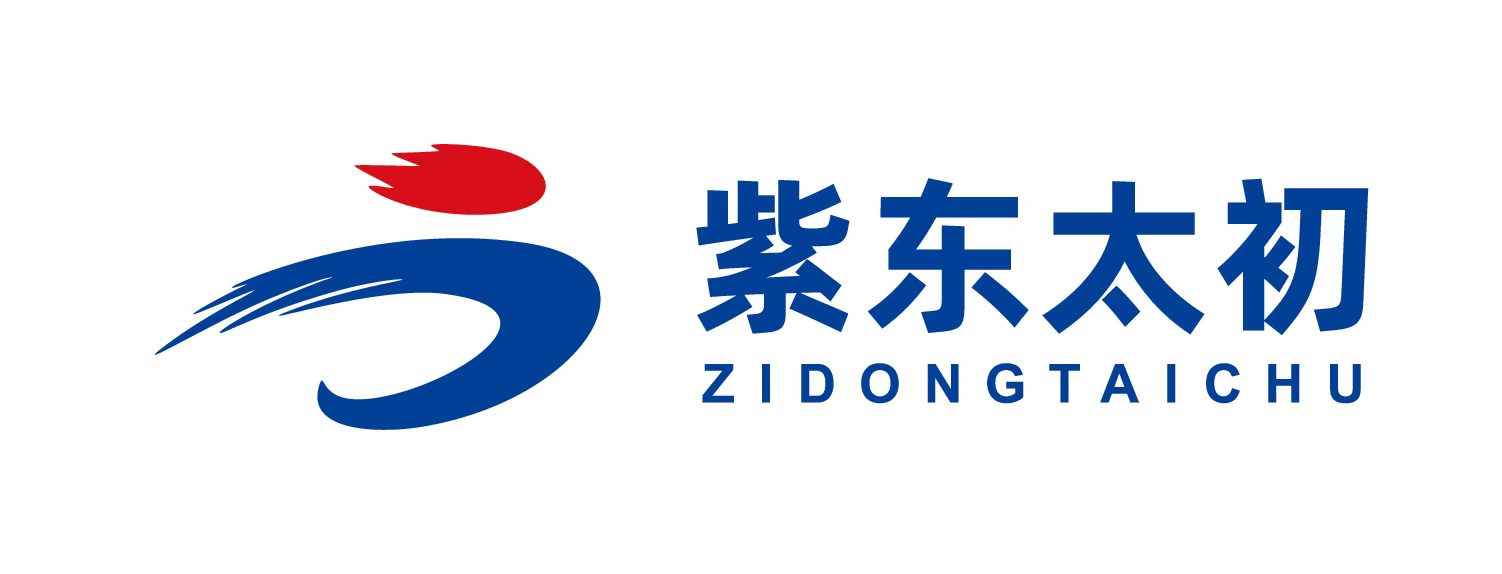}{trim=140bp 135bp 120bp 130bp,clip}\hfill
\contributor{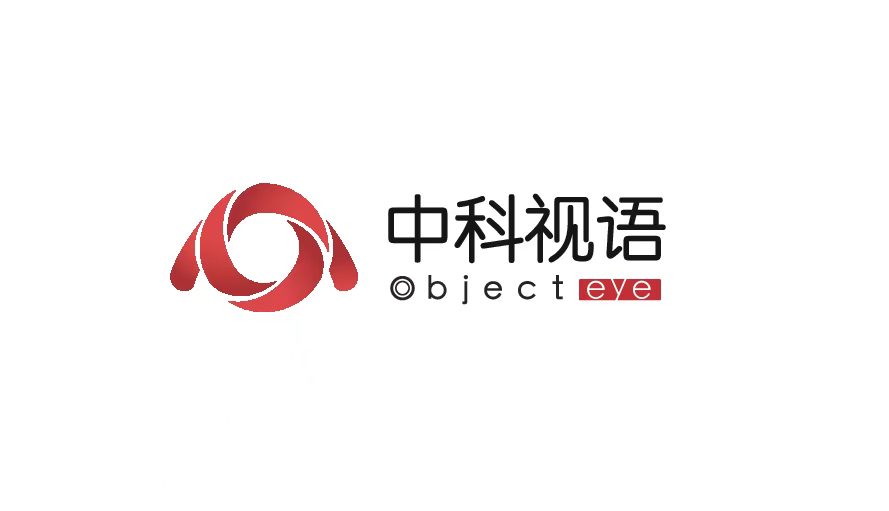}{trim=155bp 185bp 215bp 175bp,clip}
\par
\endgroup
\vspace{6pt}

\begin{abstract}
Realizing Embodied AI for Science requires robots to translate experimental goals into reliable physical procedures, reducing hazardous and repetitive human work while generating evidence for scientific discovery. Although language-based planning and learned manipulation provide complementary capabilities, their integration requires robots to adapt subsequent operations to actual execution outcomes. For example, an unsuccessful placement changes what must be done next, even when the experimental goal remains unchanged. Existing benchmarks assess planning, manipulation, and laboratory workflows through task completion and reasoning or execution metrics; however, aggregate performance leaves the connection between outcome-dependent choices and physical execution unresolved. To address this evaluation gap, we introduce \textbf{Labference}, a benchmark for evaluating how observed physical outcomes guide subsequent choices and how those choices translate into robotic laboratory actions. Its evaluation protocol is designed to test outcome-dependent behavioral adaptation through controlled evidence variations, while independently measuring operation selection, physical execution, and task completion. Matched Oracle-assisted continuations from shared checkpoints then assess which forms of support improve subsequent choices and task completion. Initial weighing evaluations reveal target-selection errors in the tested VLA policies despite successful object transfers, whereas GPT-driven control achieves correct selection but encounters execution failures and substantial computation costs. Together, these findings highlight the need to connect task-appropriate reasoning with reliable, timely action. By making this connection measurable and diagnosable, Labference supports targeted improvements toward autonomous experimentation for Embodied AI for Science.
\end{abstract}
\vspace{5pt}
\end{@twocolumnfalse}]

\section{Introduction}
\label{sec:intro}

Autonomous laboratories promise to reduce hazardous, repetitive human work and accelerate scientific exploration by connecting experimental design, physical execution, and the interpretation of results. Realizing this promise requires reliable handling of samples, reagents, and instruments, often around chemicals, heated vessels, and moving equipment where sustained precision and attention are essential. Laboratory automation can reduce hands-on exposure and repetitive workload while increasing throughput~\cite{leong2025safety,zhang2025robochemist}; the longer-term goal is a laboratory that selects experiments, carries them out, and uses the resulting evidence to guide its next investigation~\cite{tom2024selfdriving,canty2025science}. For that cycle to produce meaningful scientific evidence, the robot must preserve experimental requirements throughout the physical procedure and adapt when an operation produces an unexpected result.

Progress in experimental planning and robot control provides complementary foundations for this goal. For example, Coscientist has demonstrated synthesis planning, instrument programming, and experiment optimization with tool support~\cite{boiko2023coscientist}, while vision-language-action policies such as RT-2 map visual observations and instructions to robot actions across varied manipulation settings~\cite{brohan2023rt2}. Systems also connect these capabilities directly: SayCan grounds language-based skill selection in robotic affordances, and RoboChemist combines multimodal planning, visual targets, and execution monitoring for chemical procedures~\cite{ahn2022saycan,zhang2025robochemist}. Although these advances establish routes from high-level requirements to physical action, an attempted operation can leave the scene in a state different from the one the plan assumed. As objects move and measurements change, the robot must therefore use the observed outcome to determine what remains to be done and guide its next action accordingly.

The ability to execute a skill thus differs from the ability to apply it appropriately: the same grasp-and-insert motion can satisfy or violate a procedure depending on the selected tube, destination, and timing. Geometric grounding methods such as VoxPoser and ReKep translate language and visual evidence into spatial objectives or manipulation constraints~\cite{huang2023voxposer,huang2024rekep}; when an outcome changes the required operation, the system must also revise the objective guiding those constraints. Laboratory autonomy therefore involves two coupled feedback loops: measurements guide the next experiment, while manipulation outcomes guide the next operation within it~\cite{fakhruldeen2022archemist,dai2024autonomous}. Both depend on selecting appropriate operations and executing them while maintaining sample identity, meeting geometric tolerances, and respecting instrument timing and procedural dependencies~\cite{lan2025autobio,salazar2026adept}.

For instance, when loading two equal-mass tubes into opposite centrifuge slots, the destination of the second tube depends on where the first tube actually settles. If the first tube enters a different slot from the intended one, the second destination changes; if insertion fails, the robot must complete that placement before advancing. The same dependence appears in feedback weighing: for a 200\,g target, a stable reading of 150\,g calls for adding 50\,g, whereas a reading of 180\,g calls for 20\,g. Because a weight that misses the pan leaves the load unchanged, the next adjustment must follow the measured result rather than the attempted placement. In both cases, the robot must apply a procedural rule to an observed outcome and direct its manipulation skill to the resulting target.

Where loading and weighing use current occupancy or readouts, sample retrieval additionally depends on an earlier observed result. For example, if a colleague or another robot moves a rack after a tube has been placed, an instruction to retrieve that tube still refers to the same sample even though its world position has changed. Among visually identical tubes, successful retrieval therefore requires remembering the sample's actual local slot and locating that slot in the rack's current pose. Since grasping a neighboring tube can succeed physically while violating the procedure, retrieval evaluation must distinguish the remembered target association, its current localization, and the execution of the grasp.

Across these examples, the reasoning--action connection has a concrete meaning: commands must preserve the selected object, destination, quantity, and timing, while subsequent choices must reflect the observed result. A changed pose calls for updated motion, a failed placement calls for recovery, and confirmed progress enables the next operation. Evaluating this connection addresses a central requirement of \emph{Embodied AI for Science}: turning experimental plans into physical procedures that satisfy their requirements and produce interpretable outcomes.

\begin{figure*}[!t]
 \centering
 \includegraphics[width=\textwidth]{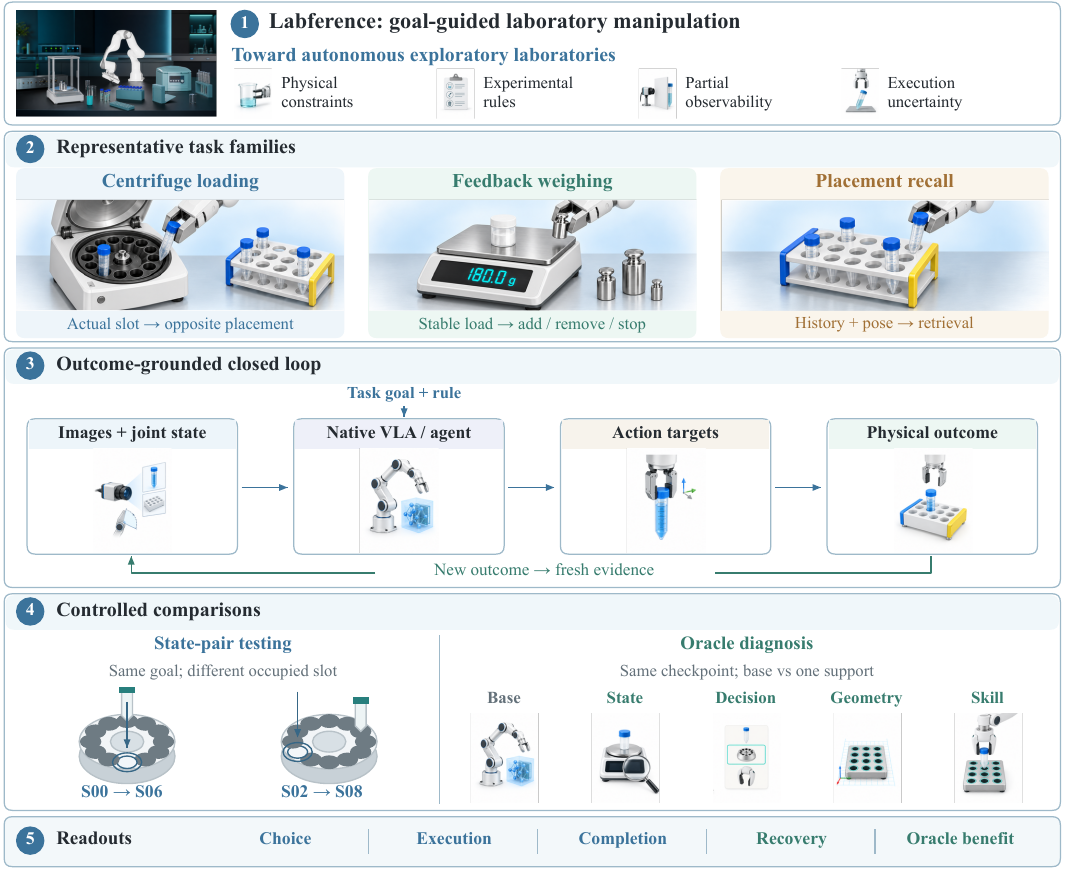}
 \caption{\textbf{Labference overview.} (1) Autonomous laboratories motivate goal-guided operational reasoning. (2) Task families expose spatial, numerical, and history-dependent choices. (3) Goals guide policies; physical outcomes generate fresh evidence. (4) State pairs test evidence-dependent choices; matched checkpoint continuations compare base systems with state, decision, geometry, or execution support. (5) Choice, execution, completion, recovery, and support benefit are reported separately. Generated illustrations depict task concepts; the weighing readout is illustrative.}
 \label{fig:state-choice}
\end{figure*}

These operational demands create a corresponding measurement problem. Chemistry3D, AutoBio, LabUtopia, Labimus, and LabDex establish foundations in scientific simulation, instrument interaction, dexterity, quantitative precision, and extended workflows~\cite{li2024chemistry3d,lan2025autobio,li2025labutopia,wu2026labimus,tang2026labdex}, while LABSHIELD examines hazard perception and safety-critical planning~\cite{sun2026labshield}. Reasoning benchmarks further study plan adaptation, history-dependent manipulation, and decision-making interfaces~\cite{tupini2026asgard,chen2026rmbench,li2024eai}. Although these evaluations characterize important aspects of laboratory competence, deployment also requires understanding how outcome interpretation, operation selection, and execution work together: does a changed physical outcome lead to the required change in choice, and does that choice guide successful manipulation? Aggregate completion scores combine wrong-target selection with failed execution, whereas isolated interface scores leave their downstream effects unresolved. Addressing this gap requires controlled changes to physical evidence and targeted support within the same procedure, so that both the adaptation of choices and their consequences can be measured.

Addressing this gap calls for three connected measurements. To test \emph{evidence use}, the same goal and rule must require different choices after different outcomes; controlled variations distinguish this adaptation from repetition of a predictable sequence~\cite{jiang2026benchmarkaudit}. Once a choice is identified, \emph{choice and execution} must be recorded separately so that selecting the wrong slot and failing to insert into the correct slot remain distinct outcomes. To make the diagnosis actionable, \emph{matched support conditions} must then test what changes when state recognition, operation selection, localization, or execution receives assistance. Common physical checkpoints connect these measurements: they fix the starting condition, identify the tested interface, and expose the consequence of support during continued manipulation.

Within this evaluation, we use \emph{state-dependent reasoning} to denote applying a task's rules to current evidence and relevant interaction history to select an operation, object, destination, or appropriate waiting and stopping condition. The observable relationship between evidence and choice therefore supplies the evaluation target, while physical confirmation establishes whether the resulting action fulfilled the requirement.

We introduce \system{}, a benchmark for studying this relationship during robotic laboratory manipulation (Figure~\ref{fig:state-choice}). Each task specifies the relevant evidence, the set of legal next operations, and the physical events that confirm progress. For example, a placement advances the task once the object is released, supported, and stable, while a weighing decision uses a stable measured load. By tying progress to these events, the formulation accommodates multiple valid operation orders and grounds subsequent decisions in actual outcomes. Controlled instance pairs then preserve the goal and task rule while varying the evidence that should change the choice; online evaluation separately records the first identifiable choice, its execution, and overall completion, together with subsequent correction and recovery. This structure accommodates spatial occupancy, measured quantities, equipment states, and sample-location history across laboratory procedures.

Once a behavioral failure has been identified, the next question is which support helps the system make progress. \system{} addresses this question through online checkpoints that measure state recognition, operation selection, target localization, or execution from common physical starting conditions. Matched Oracle continuations then alter one registered input or interface output, allowing the subsequent choice and physical result to be compared with the base branch. For instance, observed occupancy can provide state support, a legal symbolic operation can provide decision support, and candidate coordinate frames can provide geometry support; a reference executor can also attempt the operation selected by the system. Comparing these outcomes alongside checkpoint errors helps identify targeted improvements for further testing.

The framework accommodates both native vision-language-action policies and agents that combine these policies with other tools, applying compatible support at declared observation or execution interfaces. Across these system designs, evaluation asks whether behavior responds to relevant evidence, whether the selected operation is carried out successfully, and which support improves subsequent performance. The resulting measurements can guide targeted improvements to be tested on held-out instances, connecting the diagnosis of operational failures to the broader goal of sustained autonomous experimentation.

Initial feedback-weighing evaluations illustrate the value of measuring selection and execution together: the tested VLA policies sometimes transfer an incorrect standard weight, whereas the revised GPT-driven system selects the required weight in all ten tested trials but completes eight. Even after correct selection, drops affect completion and prolonged deliberation limits practical deployment. These observations motivate further evaluation of how action generation and recovery preserve the chosen object, destination, or remembered target association, and how efficiently the system can respond to new evidence.

Our contributions are twofold:
\begin{itemize}
 \item \textbf{An online benchmark for state-dependent laboratory operations.} We connect physically confirmed outcomes to legal next choices, use controlled evidence variations to test behavioral adaptation, and measure selection, execution, and task completion separately.
 \item \textbf{A diagnostic protocol connecting choices to their consequences.} Shared checkpoints and matched Oracle continuations measure accessible capabilities and the downstream benefit of targeted support, supporting systematic investigation of bottlenecks in laboratory autonomy.
\end{itemize}

\section{Related Work}
\label{sec:related}

\begin{table*}[!t]
\centering
\begingroup
\hypersetup{allcolors=black}
\caption{Benchmark comparison by evaluation design. Beyond domain and robot execution, the table distinguishes what is scored, whether decision-relevant evidence is paired, and where support enters. The \system{} row describes the protocol defined in this report.}
\label{tab:related}
\fontsize{8.5}{10.5}\selectfont
\setlength{\tabcolsep}{3pt}
\renewcommand{\arraystretch}{1.15}
\newcommand{\yes}{\ensuremath{\checkmark}}
\begin{tabularx}{\textwidth}{@{}L{0.235\textwidth}*{3}{>{\centering\arraybackslash}X}L{0.14\textwidth}>{\centering\arraybackslash}X L{0.175\textwidth}@{}}
\toprule
Benchmark & \shortstack{Lab\\tasks} & \shortstack{Robot\\execution} & \shortstack{History\\evidence} & Scored reasoning output & \shortstack{Evidence\\pairs} & Oracle / expert support site \\
\midrule
\multicolumn{7}{@{}l}{\textit{Laboratory environments and scientific procedures}} \\
Chemistry3D \cite{li2024chemistry3d} & \yes & \yes & -- & -- & -- & -- \\
AutoBio \cite{lan2025autobio} & \yes & \yes & -- & -- & -- & -- \\
LabUtopia \cite{li2025labutopia} & \yes & \yes & -- & -- & -- & -- \\
Labimus \cite{wu2026labimus} & \yes & \yes & -- & -- & -- & -- \\
LabDex \cite{tang2026labdex} & \yes & \yes & -- & -- & -- & -- \\
LABSHIELD \cite{sun2026labshield} & \yes & -- & -- & \yes\ Safety QA & -- & -- \\
\midrule
\multicolumn{7}{@{}l}{\textit{Feedback, planning, and interface-level diagnosis}} \\
LoHoRavens \cite{zhang2023lohoravens} & -- & \yes & -- & -- & -- & \yes\ Termination \\
LoTa-Bench \cite{choi2024lota} & -- & -- & -- & -- & -- & -- \\
VLABench \cite{zhang2025vlabench} & -- & \yes & -- & \yes\ Plans & -- & \yes\ Object labels \\
EmbodiedBench \cite{yang2025embodiedbench} & -- & \yes & -- & -- & -- & -- \\
Embodied Agent Interface \cite{li2024eai} & -- & -- & -- & \yes\ Module reports & -- & \yes\ Transition model \\
AsgardBench \cite{tupini2026asgard} & -- & -- & \yes & -- & -- & -- \\
\midrule
\multicolumn{7}{@{}l}{\textit{History, process understanding, and evidence acquisition}} \\
RoboVQA \cite{sermanet2023robovqa} & -- & \yes & \yes & \yes\ QA, plans & -- & \yes\ Cognitive / physical intervention \\
LIBERO-Mem \cite{chung2025liberomem} & -- & \yes & \yes & -- & -- & -- \\
RMBench \cite{chen2026rmbench} & -- & \yes & \yes & -- & -- & \yes\ Termination \\
WatchAct \cite{li2026watchact} & -- & \yes & \yes & \yes\ Plans & -- & \yes\ Oracle plan \\
RoboProcessBench \cite{xia2026roboprocess} & -- & -- & \yes & \yes\ Process QA & -- & -- \\
EFM-10 \cite{he2026efm} & -- & \yes & -- & -- & -- & -- \\
ActiveManip-Bench \cite{liu2026sapave} & -- & \yes & -- & -- & -- & -- \\
\midrule
\textbf{\system{}} & \yes & \yes & \yes & \yes\ Legal next choice; interface reports & \yes & \yes\ State, decision, geometry, skill \\
\bottomrule
\end{tabularx}
\par\vspace{3pt}
\begin{minipage}{\textwidth}
\fontsize{8}{9.5}\selectfont
\yes: explicitly described in at least one evaluation setting; --: not established in the cited protocol under these definitions.
\textit{Robot execution}: simulated or real manipulation control, rather than named simulator actions alone.
\textit{History evidence}: historical or temporal evidence is an explicit test input or ablation.
\textit{Scored reasoning output}: reports or choices evaluated separately from physical completion; QA denotes question answering.
\textit{Evidence pairs}: matched goal and rule, a controlled change in decision-relevant evidence, and joint scoring of the required choices across both members.
\textit{Support site}: the information or interface supplied during evaluation; demonstration generation alone does not qualify.
RoboVQA reports cognitive and physical intervention rates. \system{} pairs supported and base continuations and records choice changes and physical outcomes at each declared support site.
\end{minipage}
\endgroup
\end{table*}

\subsection{Autonomous laboratories and scientific manipulation}
Self-driving laboratories integrate experimental execution with data-driven experiment selection~\cite{tom2024selfdriving,canty2025science}. Coscientist uses language models, documentation retrieval, code execution, and laboratory automation to plan and perform chemical experiments~\cite{boiko2023coscientist}. Mobile robotic chemistry demonstrates exploratory workflows spanning instruments and measurements~\cite{dai2024autonomous}. RoboChemist connects a VLM planner and monitor to a VLA through image-based targets for long-horizon chemical procedures~\cite{zhang2025robochemist}. These systems illustrate several ways to turn experimental plans into action: instrument programs, coordinated robotic workflows, and visually grounded manipulation policies. \system{} studies the operational choices within this connection, asking whether actual outcomes lead to appropriate subsequent actions. Safety analyses emphasize the hazards that laboratory autonomy must address~\cite{leong2025safety}; ADePT organizes laboratory-robot capability around adaptability, dexterity, perception, and task complexity~\cite{salazar2026adept}.

Laboratory benchmarks provide domain-specific environments and measurements. Chemistry3D integrates robot interaction with reaction visualization and liquid handling~\cite{li2024chemistry3d}. AutoBio models biological instruments, reactive interfaces, and manipulation protocols~\cite{lan2025autobio}. LabUtopia supplies high-fidelity scenes and hierarchical scientific tasks~\cite{li2025labutopia}; its assets inform the environments used here. Labimus evaluates humanoid dexterity through task completion, quantitative precision, and procedural progress~\cite{wu2026labimus}. LabDex extends laboratory evaluation to real-world demonstrations and tasks organized from atomic skills to complete workflows, reporting both success and completed-skill progress~\cite{tang2026labdex}. LABSHIELD assesses hazard recognition and safety-aware planning through multiview questions~\cite{sun2026labshield}. These works establish complementary physical, procedural, and safety requirements. \system{} connects such requirements to the next choice after an observed outcome, retaining both the selected operation and its physical realization.

\subsection{Feedback-dependent planning and diagnostic evaluation}
Language-to-action grounding takes both modular and end-to-end forms. SayCan combines the relevance of a skill to an instruction with its estimated feasibility in the current state~\cite{ahn2022saycan}. RT-2 trains vision-language models to produce low-level actions from images and instructions~\cite{brohan2023rt2}. VoxPoser composes spatial value maps for motion planning, while ReKep expresses manipulation requirements as relations among tracked visual keypoints and replans as the scene changes~\cite{huang2023voxposer,huang2024rekep}. Inner Monologue and CLIER revise plans using observations and interaction outcomes~\cite{huang2023innermonologue,nazarczuk2024clier}. These methods establish both instruction grounding and reactive constraint satisfaction. \system{} measures whether the selected object, destination, and next operation satisfy the experimental requirement after actual outcomes, and how localization or execution support changes the continuation.

Several benchmarks measure aspects of this dependence. LoHoRavens tests long-horizon relational and arithmetic manipulation with alternative observation-feedback interfaces~\cite{zhang2023lohoravens}. LoTa-Bench compares language-oriented planners by executing their plans in household simulators~\cite{choi2024lota}. VLABench measures reasoning-related plans and physical execution, including conditions with object-label support~\cite{zhang2025vlabench}. EmbodiedBench organizes visual-agent evaluation into capability subsets spanning perception, spatial understanding, instruction following, and long-horizon planning~\cite{yang2025embodiedbench}.

Two particularly close works clarify how adaptation and diagnosis are evaluated. AsgardBench constructs observation-dependent branches under a common instruction and compares visual and action-feedback conditions, isolating interactive planning through high-level simulator actions~\cite{tupini2026asgard}. Embodied Agent Interface instead standardizes goal interpretation, subgoal decomposition, action sequencing, and transition-modeling outputs, including error analysis and sensitivity to predicted transition models~\cite{li2024eai}. The former emphasizes adapting a plan to feedback, while the latter exposes the correctness of intermediate interfaces. For contact-rich laboratory operation, the additional question is how such choices and interface outputs affect the next physical interaction. \system{} tests that transition by deriving decision opportunities from measured outcomes and comparing base and supported continuations from common checkpoints. Choice, execution, and final completion retain separate scores throughout this comparison.

\subsection{History, process understanding, and access to evidence}
RoboVQA evaluates video-grounded questions and medium-horizon plans, including live robot execution with human intervention~\cite{sermanet2023robovqa}. By distinguishing cognitive from physical intervention rates, it separates high-level corrections from low-level motor assistance. \system{} shares this interest in the relationship between planning and execution, but centers its comparison on controlled changes in the evidence that determines the next legal operation. Paired instances preserve the goal and rule while changing that evidence; common checkpoints then compare base continuations with state, decision, geometry, or skill support. This links two questions within the same physical procedure: whether choices track the changed evidence and which assistance improves their realization.

LIBERO-Mem tests object-specific histories under partial observability~\cite{chung2025liberomem}, while RMBench studies memory-dependent manipulation and the effect of memory components and GT subtask termination~\cite{chen2026rmbench}. WatchAct separates reasoning from observed human behavior, execution given an Oracle plan, and integrated performance~\cite{li2026watchact}. RoboProcessBench evaluates understanding of recorded manipulation progress, contacts, temporal relations, and subsequent primitives~\cite{xia2026roboprocess}. Together, these benchmarks connect historical evidence to plans and process judgments. Our placement-recall family couples the robot's own realized placement with subsequent rack relocation. In the same-appearance condition, the target association comes from the placement history, while the current rack pose determines where to act. Recall reports, first retrieval choices, and retrieval execution are recorded separately; the evidence and history delivered to each system define the memory condition.

Historical and current evidence both depend on how observations are acquired. EFM-10 evaluates exploratory and focused manipulation with bimanual active perception and force sensing~\cite{he2026efm}, while ActiveManip-Bench, introduced with SaPaVe, evaluates manipulation under changing viewpoints and occlusion~\cite{liu2026sapave}. Since these conditions change what the policy can observe, sensor access and observation timing must accompany any comparison of its choices. \system{} therefore records evidence timing, authorized history, and observation conditions when matching decisions and Oracle continuations.

\subsection{Policy systems, infrastructure, and the comparison boundary}
Harness VLA composes a VLA with planning, memory, grounding, and motion tools~\cite{zhang2026harnessvla}; this composition is a tested policy system. The VLA Evaluation Harness provides infrastructure for model--simulator exchange~\cite{choi2026vlaeval}. Their roles differ from the task and diagnostic benchmarks in Table~\ref{tab:related}. \system{} uses declared policy interfaces to test evidence-dependent behavior and support at compatible intervention sites. Native policies and tool-using agents retain their observation, action, and budget conditions in each comparison.

The resulting emphasis is the connection among three measurements: adaptation to changed operational evidence, correctness of accessible interfaces at common checkpoints, and downstream response to targeted support. Each Oracle condition supplies or replaces a specified input or output while the continuation records actual choices and physical outcomes. Benchmark audits of shortcuts and evaluation dependence further motivate controlled evidence changes and fixed comparison populations~\cite{jiang2026benchmarkaudit}. This combination makes laboratory reasoning measurable within the manipulation process and links diagnosis to candidate improvements that can be tested on held-out procedures.

\section{Evaluating State-Dependent Operations}
\label{sec:definition}

\subsection{From action sequences to outcome-dependent choices}
The central question is whether a robot chooses its next operation from what actually happened. A planned placement and a confirmed placement can require different responses, even under the same instruction. For a task version $\tau$, the evaluator maintains a task state $z_t$, a goal $g$, and a legal next-operation set $\legal_\tau(z_t,g)$. The state contains confirmed object relations, completed obligations, and relevant device or stability timers. An operation $u=(k,b,r)$ specifies its type, object, and destination; Wait and Stop use empty object and destination fields where appropriate. Low-level robot commands are denoted by $a_t$ and remain distinct from these task-level operations.

Progress follows physical evidence:
\begin{equation}
 z_{t+1}=F_\tau(z_t,e_{t+1},\Delta t),
 \label{eq:update}
\end{equation}
where $e_{t+1}$ contains outcomes derived from measurements of the physical scene state $x_{t+1}$ and $\Delta t$ updates device processes. The set $\legal_\tau(z_t,g)$ is the reference against which a policy's choice is tested; a policy can also select an operation outside this set. Several operations may be legal, allowing alternative orders and recovery. Rule application is therefore evaluated by membership in a state-dependent set.

\subsection{Task contracts}
To apply this state-update rule consistently across tasks, a TaskContract specifies six connected elements: decision-relevant state, authorized evidence, legal operations, physical confirmation predicates, waiting and terminal conditions, and supported diagnostic interfaces. Protocol dependencies determine when an operation becomes enabled. Retry and wait transitions operate over the evolving state, while a dependency graph describes ordering constraints among confirmed obligations. Because progress depends on confirmed events, each event retains its simulation time, object identity, measured target, and associated attempt. These identifiers prevent repeated contact or repeated requests from completing the same obligation multiple times.

The task state is evaluator data. The policy receives an observation $o_t$ containing registered camera images, robot state, the goal and rule, and its authorized history. A contract links each scored decision to evidence available through this interface. Current visibility and historical delivery are checked separately: an earlier observed placement can support later retrieval even when its identity is presently ambiguous. Checkpoint populations fix the evidence needed to distinguish the legal alternatives.

\subsection{Shared task-design principles}
Three principles connect the laboratory settings. First, \emph{outcomes determine remaining work}: an unsupported placement leaves its obligation unfinished. Second, \emph{evidence determines when a decision is scorable}: the registered observations and history must expose the facts needed for the choice. Contracts specify a supported inspection, waiting, or retry operation for recoverable uncertainty. Third, \emph{physical confirmation is independent of policy intent}: release, support, alignment, and continuous stability establish completion under versioned task-specific tolerances.

Spatial loading and feedback weighing illustrate this design. An occupied centrifuge slot determines the opposite destination; a measured load determines which weight to add or remove. Basic conditions provide the initial state, while extended conditions require the robot to create it and respond to its actual outcome. These settings share the same decision model. Initialized objects establish context; policy credit begins with the subsequent interaction.

History-dependent retrieval uses the same event model. A released placement binds a sample to its actual slot. A placement deviation and the subsequent retrieval reference can therefore coexist: the former records how execution differed from the intended target, while the latter follows the location the policy could observe. Spatial loading, quantitative feedback, and remembered placement are complementary applications of the outcome-to-choice relation.

\subsection{Controlled changes to the evidence}
The event model specifies the appropriate choice in each state; controlled instances test whether a policy responds when that state changes. Matched instances preserve the instruction and nuisance conditions while changing a fact that requires a different next operation. An answer-changing pair is registered at a choice where the two reference sets are disjoint. For example, a centrifuge comparison holds the second tube ready for insertion and varies the first tube's actual slot, making the destination of the already-held tube the scored choice. Contrasting successful and failed placements instead tests recovery selection. Stable versus unsettled readings extend the same principle to temporal decisions for interfaces that expose Wait or Stop.

Task-state, scene-layout, and policy seeds are stored separately. The generator samples target quantities, occupancy, distractors, and initial progress independently of nuisance placements and appearances, subject to physical feasibility constraints. A pair shares a nuisance draw and records its realized layout hash. The designated state variable and its necessary observable consequences are then varied together: changing a supported load also changes the displayed reading. Physical initializations and action prefixes are verified before the scored handoff.

Geometric transformations provide a complementary control: the correct local target remains unchanged while its world position and axis move. Answer-changing pairs therefore test operation selection, whereas geometric-equivariance pairs test whether the same local choice can be realized after relocation. The geometry-preserving controls in Figure~\ref{fig:state-choice} preserve object-relative relations while changing world coordinates. Both pair types retain a fixed instance list and all failed attempts.

\section{Laboratory Task Families}
\label{sec:families}

The state-dependent formulation becomes concrete in three laboratory settings (Figure~\ref{fig:state-choice}). Centrifuge loading creates a spatial constraint, weighing changes a numerical residual, and rack placement creates a location that must be recovered after an interruption. Although the relevant evidence differs, each setting makes a realized outcome determine the next appropriate operation. The resulting tasks examine spatial, numerical, and historical dependencies through complete interactions; Appendix~\ref{app:original-catalog} describes the broader L01--L12 and R01--R12 catalog.

\subsection{Centrifuge loading: acting on actual occupancy}
An equal-mass tube pair must occupy opposite rotor slots. The policy observes the rotor and tubes, grasps a tube, selects a destination, and completes insertion and release. For a rotor with an even number $n$ of slots, zero-based local index $j$ has opposite index $(j+n/2)\bmod n$. The reference environment uses twelve calibrated slots. Rotor orientation changes their world coordinates while preserving this local relation.

\paragraph{Basic and extended conditions.}
In the basic condition, one tube is already stably seated and the second is available for grasping. The next destination follows from the occupied slot. In the extended condition, both tubes start outside the rotor. The first can occupy any legal empty slot with an available opposite. Its actual stable insertion establishes the reference for the second. A failed grasp leaves the first tube available for another attempt; a shallow or unreleased insertion leaves its placement unfinished. A stable insertion in another legal slot creates a new, observable spatial constraint. The record retains both the intended first slot and the resulting occupancy.

Consider a policy that intended to seat the first tube at local slot 1. If it actually settles at slot 2, the next destination is slot 8, whereas successful seating at slot 1 calls for slot 7. If the first placement fails, the pending operation concerns that tube. These branches distinguish following a planned sequence from responding to what the sequence achieved. A controlled destination test holds the second tube ready and changes the first tube's actual occupied slot; a separate failed-placement checkpoint tests recovery selection.

\paragraph{Choice and physical confirmation.}
The operation vocabulary is Pick, Insert, Wait, and Stop, with tube and slot arguments where applicable. The insertion recognizer records the first directed attempt toward a slot. The physical predicate subsequently checks tube-axis alignment, radial clearance, depth, support, release, and stability. Choosing the opposite slot and successfully seating the tube therefore produce separate measurements. Final completion requires both tubes to remain in the opposite configuration for the terminal stability window. The setting evaluates equal-mass loading and the decisions that establish it.

\subsection{Feedback weighing: revising a numerical plan}
Where centrifuge loading derives the next destination from occupancy, feedback weighing derives the next adjustment from a measured quantity. The robot must bring a loaded balance to a stated target using labeled standards of $10$, $20$, $50$, and $100$\,g, one of each. An unlabeled sample is already supported on the pan and stays there throughout the task. The robot observes the measured display, stability indicator, and current standard-weight arrangement, then adds a standard, removes one, waits for a valid reading, or finishes.

\paragraph{Basic and extended conditions.}
The basic condition starts with all standards on the table. Targets are sampled from $160$, $180$, and $200$\,g; the sample mass is chosen so that exactly one standard completes the target. For example, under a $180$\,g instruction, a stable $130$\,g reading calls for the $50$\,g standard, while $160$\,g calls for $20$\,g. Matching appearance and layout while changing the hidden preload creates an evidence-dependent choice with the same instruction.

The extended condition starts with a subset of standards on the pan. Targets are sampled from $200$, $220$, and $240$\,g, and the desired final subset contains two or three standards. The initial and desired subsets differ by at least two additions or removals. If $W_t$ is the supported subset and $W^*$ the goal subset, enabled corrections remove members of $W_t\setminus W^*$ or add members of $W^*\setminus W_t$. Several orders can therefore be correct. The evaluator stores $W^*$; the policy derives the required changes from the instruction, readout, and visible standards.

For instance, with a $100$\,g sample, a $220$\,g target, and a $50$\,g standard initially on the pan, the required final subset is $100+20$\,g. Removing $50$\,g and adding the two required standards can proceed in several orders. After every release, actual support and the new stable reading establish what remains to be done. If a weight misses the pan, the residual obligation remains. If a disturbance resets stability, the next eligible measurement occurs after the new stable window.

\paragraph{Measurement and completion.}
The display is driven by the pan's measured support force, with empty-pan calibration, filtering, and a visible stability indication. The sample mass remains evaluator-side data. Final success combines the target reading, exact supported standard set, release, low motion, and continuous stability. Thus a temporarily correct reading during hand pressure and a correctly released load have different physical outcomes. Numerical choice, loading execution, timing, and completion are recorded independently. Section~\ref{sec:implementation} specifies the sensor and window parameters.

\subsection{Placement recall: using an earlier observed result}
Placement recall extends outcome dependence to evidence acquired earlier in the interaction. A robot places a tube in a freely chosen empty hole of a $3\times4$ rectangular rack, withdraws, and later retrieves that same tube after occlusion. Its actual released slot creates the reference for retrieval, so later localization must preserve the sample--slot association established during placement. To support this localization after movement, the rack carries A1--C4 labels, asymmetric long- and short-edge colors, and an A1 orientation marker.

\paragraph{Basic and extended conditions.}
The basic condition uses an empty rack, one tube, and a stationary occluded interval. It provides a grasp, placement, and retrieval reference. The extended condition adds three distractor tubes and changes the rack pose during occlusion. In the same-cap condition, all four tubes share the same visible appearance; the policy can associate its earlier placement with a local slot and locate that slot in the current rack frame. In the distinct-cap condition, all caps have different colors, with target color balanced independently of slot and movement. This provides an additional appearance cue. Comparing these conditions measures behavior under different available evidence.

For example, a tube observed settling at B3 remains associated with B3 after a $180^\circ$ rack rotation, although its world location changes. A retrieval directed to the old world position reveals a different behavioral error from reporting the wrong local label. If the initial placement instead settles at B2, B2 becomes the retrieval reference. The record preserves the placement deviation and evaluates the later choice against the observed outcome.

\paragraph{History, selection, and retrieval.}
The single-delivery history protocol provides placement frames when the interaction occurs. After occlusion, the environment delivers fresh front and wrist images; earlier evidence can persist in the tested system's declared model state, agent context, or memory component. Episode reset clears that state. Same-cap retrieval consequently relates earlier evidence to the transformed scene, while the basic one-tube condition also permits identifying the unique visible tube.

The primary native metric uses the first deliberate pre-contact approach to a unique slot. Explicit slot reports and symbolic retrieval commitments form separately registered tracks. Correct slot selection is retained even if the subsequent grasp fails, and later correction preserves the original choice record. Physical completion requires retrieving the target, releasing it onto the marked tabletop region, and maintaining stable table support for three seconds. Distractors retain their local slots in the extended condition.

Autonomous episodes report the slots actually chosen and the fraction reaching the retrieval opportunity. Balanced checkpoints use common placement histories over all twelve slots, with a fixed denominator. The two populations answer complementary questions: how a policy manages its own interaction history, and how systems compare when they receive the same history. The occluded environment transition and its physical checks are specified in Section~\ref{sec:implementation}.

\section{Online Choice and Execution Evaluation}
\label{sec:online}

\begin{figure*}[t]
 \centering
 \includegraphics[width=\textwidth]{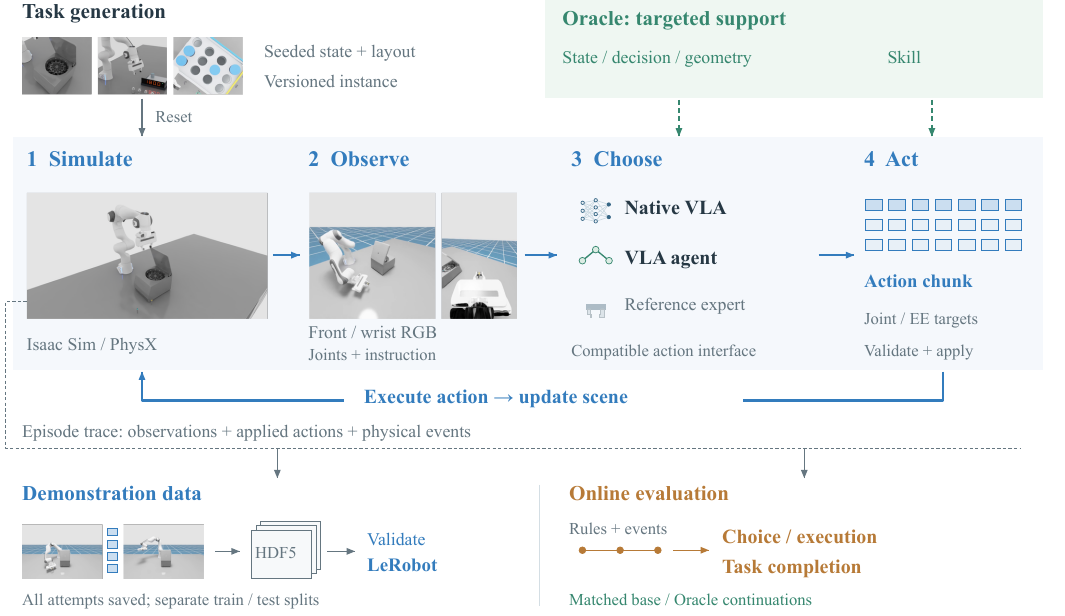}
 \caption{\textbf{Closed-loop evaluation protocol.} Actual outcomes produce fresh observations and the next choice. Oracle support enters compatible interfaces; evaluation and demonstration recording form separate outputs.}
 \label{fig:online-loop}
\end{figure*}

\begin{table*}[t]
\centering\small
\caption{Diagnostic questions and corresponding Oracle support. Direct measurements test accessible interfaces; supported continuations measure downstream benefit.}
\label{tab:diagnostic}
\label{tab:oracle-sites}
\setlength{\tabcolsep}{5pt}\renewcommand{\arraystretch}{1.12}
\begin{tabularx}{\textwidth}{@{}YYY@{}}
\toprule
Question & Direct measurement & Scoped support \\
\midrule
Was the relevant state recognized? & Visible identities, occupancy, readings and stability & Semantic: authorized observable facts \\
Was the next operation appropriate? & Membership in the legal operation set & Decision: one legal symbolic operation \\
Was the selected target localized? & Position and axis or placement-region error & Geometry: candidate frames, or selected-target localization \\
Could the chosen operation be executed? & Grasp, placement, release and stable support & Skill: a reference attempt at the selected operation \\
\bottomrule
\end{tabularx}
\end{table*}

\subsection{Policy interfaces and information boundaries}
The task families define what constitutes appropriate progress; the online protocol connects that reference to the evidence and actions of a tested system. A native VLA receives images, robot state, and instructions and returns actions, while an agent may additionally retain context and invoke declared tools. Both act in the simulator and receive observations of the resulting state (Fig.~\ref{fig:online-loop}). Their behavior therefore depends on the model revision, memory configuration, tools, observation channels, and action adapter, which jointly define the tested system. An evaluation harness supplies transport and episode scheduling; any option that changes observations or applied actions becomes part of the recorded system condition.

Each policy profile declares an \emph{ActionSpec}: joint or end-effector representation, dimensions, absolute or incremental mode, coordinate frame, units, gripper convention, control frequency, and maximum chunk length. The reference eight-dimensional joint interface carries seven arm targets and one profile-defined gripper command. Episode reset clears queued actions and resets the declared policy session. Episode-local memory then persists according to the selected profile. RGB and RGB-D are distinct observation conditions.

The runtime separates three data views. The \emph{policy observation} is a whitelist projection of images, robot state, public instructions, and authorized history. The \emph{Oracle context} carries a separately registered intervention. The \emph{evaluator state} contains physical references, hidden identities or masses, legal sets, and success predicates. Input checks operate on the serialized request and its final camera and prompt mapping. Delivery receipts identify the evidence that reached the policy boundary; later behavior determines how that evidence was used.

\subsection{A timed closed loop}
To associate each action with the evidence available when it was generated, one runtime component owns simulator stepping. At a control boundary it obtains the registered observations, queries the policy, validates the returned action chunk, and applies commands under the declared hold interval. Physical measurements are updated after each simulation step. Camera acquisition and delivery times are recorded alongside those measurements, linking a rendered frame to the physical event it depicts. Agent calls for observation, VLA inference, ordinary motion, and waiting use the same step owner and therefore share this timing reference.

The control rate and physics rate are separate quantities. A command held over several physics steps contributes one policy action and several physical measurements. A new evidence handoff or intervention invalidates the queued commands tied to an earlier planning boundary. Logs preserve requested and applied actions, making clipping, rejection, and buffering visible. Waiting consumes simulated time; inference latency and tool calls are additional recorded costs. In the synchronous reference runtime, physics advances during action execution and explicit waiting, and pauses during inference. Simulation-time stability and wall-clock responsiveness are consequently distinct measurements. Each response budget specifies its clock and the observation-delivery boundary from which it is measured.

\subsection{Event-defined decision opportunities}

This timing reference determines when a choice becomes scorable. An opportunity opens after relevant new evidence reaches the policy, linking an opportunity identifier, physical reference state, event time, delivered observation identifier, response window, and policy session. Initial readiness, a confirmed state change, or an observable failed attempt can trigger a new opportunity. Repeated queries of unchanged evidence retain the current identifier because they expose the same pending choice. The reference state precedes the scored interaction: a grasp is judged using the state before contact, and its confirmed result can subsequently enable an insertion opportunity.

An event recognizer identifies a behavioral choice from a versioned interaction pattern. Bilateral finger contact and closure can identify a selected object; directed entry into a unique socket can identify an insertion target. The rack-retrieval variant uses a sustained, descending approach into one local slot region before contact. Each recognizer has its own declared thresholds, independent of the stricter physical success predicates. The first identifiable choice is frozen, and revisions retain their own timestamps. A correct target followed by a dropped object receives choice credit and an execution failure.

Explicit operation reports form a separate scoring track. Their first commitment can precede motion, and agreement with subsequent behavior is measured independently. Native inactivity leaves Wait and Stop unreported; prerequisite violations and continuous terminal dwell provide the temporal measurements for action-only policies. Actions continue during the terminal window, and a disturbed terminal state restarts its timer.

\subsection{Results at three levels}
The choice level asks whether the first selected operation belongs to the current legal set. The execution level measures a physical subgoal from a common start with the same correct target and geometry. The full-task level combines the achieved terminal state with required protocol and safety conditions. Premature advancement, unnecessary repetition, early termination, and subsequent recovery remain explicit event categories. Section~\ref{sec:measurement} defines the populations and denominators used for these results.

On-policy trajectories reach different states. Their decision rates are accompanied by reach rates and opportunity counts; cross-system comparisons of the same decision use shared checkpoints. These results connect local choice quality to task completion while retaining the measurement at each level.

\section{Diagnosing Capability Bottlenecks}
\label{sec:diagnostics}

\subsection{What can the system do at a common state?}
Because complete rollouts reach policy-dependent states, comparing capabilities requires a common starting point. Verified online checkpoints hold the physical start and authorized evidence fixed by reconstructing the registered initialization and, where required, an action--observation prefix. Pose, occupancy, release, and stability are verified at the handoff. History is part of this comparison: a failed-attempt prefix and an initially unplaced object expose different evidence even when their current poses coincide. Stateful policies therefore start from a reset session and receive the same authorized prefix, or use a declared, verified session snapshot. These procedures establish the physical and informational conditions under which interface measurements can be compared.

Four tests address accessible interfaces. State recognition measures visible identities and occupancy, or displayed mass and stability. Decision testing supplies the same correct standard state and rule, then scores the first operation tuple against the legal set. Geometry testing fixes the selected target and measures its predicted position and directed axis. Execution testing supplies a correct subgoal and standard geometry, then observes physical grasp, placement, release, and stable support. Each compatible test has an online continuation through the declared controller.

For a predicted position $\hat p$ and unit directed axis $\hat v$, geometric errors are
\begin{equation}
 e_p=\|\hat p-p^*\|_2,\qquad
 e_v=\arccos\!\bigl(\operatorname{clip}(\hat v^\top v^*,-1,1)\bigr).
 \label{eq:geometry-error}
\end{equation}
Positions use the registered metric frame. Both reference and predicted axes are normalized; zero or non-finite predictions are invalid outputs. Axis error distinguishes opposing insertion directions. Placement-region tasks instead use distance to their admissible region. Semantic outputs retain field-level errors, such as occupancy correctness and absolute readout error. A report-capable adapter is part of the tested system. Action-only policies enter native behavior and compatible execution tests; their available interfaces determine the diagnostic population.

\subsection{Which support improves subsequent behavior?}
Knowing that an interface produces an error raises a further question: would support at that interface improve subsequent progress? Oracle continuations test this by comparing the base system with a supported branch from the same checkpoint. The intervention takes one of two forms. \emph{Assist} adds information while the system produces its own output, whereas \emph{Replace} substitutes an output at an existing interface and evaluates the downstream response. When the original output is accessible, it remains available for direct scoring alongside the value actually executed. The comparison thus preserves the distinction between original interface correctness and performance under replacement.

The scopes in Table~\ref{tab:oracle-sites} isolate useful contrasts. Semantic supplies facts certified by the authorized observations: visible occupancy, public identities, or a displayed load and stability indicator. Hidden sample mass and target selection remain evaluator-side quantities. Decision provides one legal operation as an explicit answer-support condition, with symbolic identifiers and no motion coordinates. When several operations are legal, a registered selection rule chooses the supplied member and the full legal set remains the scoring reference.

Geometry assistance supplies all authorized candidate frames. Geometry replacement preserves the already selected target identity and corrects that target's localization. Skill replacement uses a reference executor to attempt the selected operation, including a wrong but physically feasible object or destination. The executor's own success and failure are recorded. It defines an execution-supported system condition and requires prior validation for the operations it accepts.

End-to-end policies receive assistance through compatible input channels. Modular systems can additionally expose state-report, operation-commitment, geometry-report, and executor hooks. Intervention mode, interface site, and transport are separate fields: structured data, fixed text serialization, and tool responses can carry assistance, whereas a replacement hook identifies the output being substituted. Capability negotiation determines which combinations a service accepts.

\subsection{Intervention timing and delivery}
Support is useful relative to a particular state, so each intervention is bound to one opportunity, evidence version, and application site. The task-owned validator checks content scope before delivery. Its record stores payload identity, original and substituted outputs where available, delivery status, and the first subsequent choice. When the task state changes, support is refreshed at the next registered handoff, and actions computed before that handoff are invalidated according to the adapter's replanning contract.

Delivery, recommendation agreement, and benefit are measured separately. A receipt establishes delivery to the model or tool interface. Recommendation agreement records whether the subsequent reported or behavioral choice matches the supplied operation. The matched base branch establishes whether that choice changes with support. Benefit is the paired change in the chosen outcome measure. An unsupported interface receives an explicit compatibility status; malformed outputs on a supported interface are scored as interface failures, while transport and service faults are recorded as evaluation-validity events.

\subsection{Matched online continuations}
To attribute a performance difference to the registered support condition, base and supported branches share a verified checkpoint, tested system, sensors, ordinary tool inventory, budget convention, and reset protocol. Skill conditions additionally identify the replacement executor and its cost. The record then links delivered support, recommendation agreement or substitution, the resulting choice, and the physical outcome. Full-rollout interventions start from a matched initialization but can reach different later states as earlier support changes behavior. They are therefore reported separately from common-checkpoint continuations.

For a higher-is-better outcome $Y$, Oracle $o$, mode $m$, and a complete population of $N$ preregistered compatible pairs,
\begin{equation}
 \Delta_{m,o}(Y)=\frac{1}{N}\sum_{i=1}^{N}
 \left(Y_i^{(m,o)}-Y_i^{(0)}\right).
 \label{eq:oracle-gain}
\end{equation}
This gain measures benefit from the specified support. Error metrics use the reverse difference so that a positive value consistently denotes improvement. Checkpoint errors, recommendation agreement, and changes in physical outcomes provide complementary diagnostic evidence. A supported rollout can improve through several interacting capabilities; the intervention identifies a useful support boundary in the complete system. For example, accurate readout reporting together with improved selection under Decision support points to operation selection as a useful intervention site. Accurate selection together with improved completion under Skill support directs attention to its physical realization. These joint patterns connect interface measurements to actionable system changes.

Assist and replace comparisons match factual scope and application time. For joint supports $o_1,o_2$, the contrast $\overline Y^{o_1,o_2}-\overline Y^{o_1}-\overline Y^{o_2}+\overline Y^{0}$ measures departure from additive gains under a matched design. It is an interaction of support conditions. Candidate interventions are selected on development instances, then tested with fixed choices and budgets on an independent population.

\section{Comparison and Reporting Principles}
\label{sec:measurement}

\subsection{Choice correctness, identification, and reach}
The protocol produces identifiable choices, physical outcomes, and supported continuations. Their comparison requires explicit populations, since a failure to reach a choice differs from an incorrect choice after evidence delivery. Let $\mathcal J$ be a fixed population of verified, evidence-delivered checkpoints at which the native track has a scorable physical choice. For first recognizable choices $b_i$ and legal reference sets $\legal_i$, define
\begin{align}
 B_{\rm next}&=\frac{1}{|\mathcal J|}\sum_{i\in\mathcal J}\ind[b_i\in\legal_i],\label{eq:bnext}\\
 C_{\rm id}&=\frac{1}{|\mathcal J|}\sum_{i\in\mathcal J}\ind[b_i\ne\bot],
 \label{eq:coverage}
\end{align}
where $\bot$ denotes no identifiable choice by the response deadline. Both metrics use the same population. Accuracy among identified choices equals $B_{\rm next}/C_{\rm id}$ when $C_{\rm id}>0$ and is reported with its coverage. Rates with a zero denominator are unavailable. Explicit reports use a separate population and the first reported operation in place of $b_i$; Wait and Stop are scored there when applicable. Checkpoint preparation failures remain in the registered instance list and its validity report; full-population comparisons require all registered starts to pass preparation.

Complete rollouts generate their own opportunity sequences as execution proceeds. Their legal-choice rate describes the states reached by that policy and is accompanied by episode and opportunity counts. For a registered later decision, let $N_{\rm run}$ be the fixed rollout count, $N_{\rm reach}$ the runs that reach it with delivered evidence, and $N_{\rm correct}$ the runs whose first choice is correct. Report $N_{\rm reach}/N_{\rm run}$, $N_{\rm correct}/N_{\rm reach}$, and $N_{\rm correct}/N_{\rm run}$ together. This distinguishes failure to reach a decision from failure to choose correctly after reaching it. Identification is additionally reported among reached opportunities.

\subsection{Execution, completion, and recovery}
Choice correctness establishes whether the selected operation is appropriate. To measure its physical realization under matched requirements, fixed-subgoal execution uses a common population $\mathcal K$ of verified starts with the same correct operations and geometric references:
\begin{equation}
 E_{\rm fixed}=\frac{1}{|\mathcal K|}\sum_{i\in\mathcal K}
 \ind[\text{physical subgoal confirmed in }i].
 \label{eq:execution}
\end{equation}
Execution conditioned on each policy's own correct choices is a complementary on-policy statistic; the systems can reach different subsets of starting states.

A terminal physical predicate $P_\tau(x_t)$ combines task-specific placement, release, support, precision, and motion requirements. Its dwell counter advances only while the conjunction holds and resets when any requirement fails. Physical completion requires this counter to reach the registered duration. Protocol and safety checks are then reported alongside physical completion and determine valid task completion. Sensor and log validity are prerequisites for interpreting these measurements.

Clean completion additionally requires correct identified choices at all eligible opportunities and the absence of recorded operation failures or protocol violations. Recovered completion requires a documented recoverable error or operation failure, its resolution, and valid task completion. An irreversible protocol or safety violation continues to invalidate completion even if the final physical arrangement is restored. Earlier incorrect choices remain in their own records. Physical completion with unidentified choices is reported with incomplete choice coverage. Clean and recovered completion thus describe verified trajectory histories, while the component reports retain their own units and denominators.

\subsection{State-changing and geometric pairs}
Single-instance correctness can coincide with a persistent target preference. State-pair performance tests adaptation by requiring legal choices in both members of a pair with different required answers:
\begin{equation}
 P_{\rm state}=\frac{1}{|\mathcal B|}
 \sum_{(i,i')\in\mathcal B}
 \ind[b_i\in\legal_i]\ind[b_{i'}\in\legal_{i'}].
 \label{eq:pair}
\end{equation}
Here $\mathcal B$ is the fixed list of answer-changing pairs defined in Section~\ref{sec:definition}; its legal sets are disjoint at the scored choice. The explicit-report track substitutes its own reported choices and applicability conditions. Geometric controls preserve local target identity. A rigid transform $(R,d)$ maps the reference mouth and directed axis as $p'=Rp+d$ and $v'=Rv$; localization errors and physical execution are evaluated in the corresponding transformed scene.

\subsection{Compatibility, incomplete data, and uncertainty}
Capability preflight defines the population eligible for each interface condition before evaluation. Unsupported conditions receive a compatibility result. Within a registered population, missing runs and service faults retain their identities and reduce completion coverage; ordinary policy failures remain measured outcomes. If $s$ pairs are verified correct and $u$ pairs remain unresolved among $N$ planned pairs, provisional paired correctness is bounded by $s/N$ and $(s+u)/N$. Such bounds describe incomplete evidence. Equation~\eqref{eq:oracle-gain} is reported as a full-population gain after every registered pair has valid outcomes.

Result identities include task and evaluation versions, asset and layout revisions, tested system, observation condition, action convention, history protocol, and intervention mode and site. These identities are checked before aggregation. An agent built around a VLA changes tools, context, and computation; its comparison with the underlying VLA is therefore a system-level comparison. Within-system Oracle contrasts retain the declared sensor and budget conditions. Task aggregates use a fixed task list and expose per-task coverage.

Uncertainty is estimated over episodes or matched pairs \cite{agarwal2021precipice}. All decisions from one rollout stay together during resampling, and matched branches stay together during pair-level resampling. Online results and replay audits share the physical-event scorer. Together, choice, execution, completion, and support response characterize state-dependent manipulation at complementary levels.

\section{Reference Implementation and Reproducibility}
\label{sec:implementation}

\subsection{Shared runtime and task-owned measurements}
The preceding measurements depend on a consistent link between physical state, delivered observations, and applied actions. The reference implementation maintains this link in Isaac Sim 5.1 and PhysX with a Franka Panda manipulator and registered front and wrist cameras. Each task module owns its physical measurements, legal-operation generator, and scorer, while shared modules provide policy profiles, capability negotiation, Oracle dispatch, and episode transport. This separation allows centrifuge loading, feedback weighing, and placement recall to use the same contract structure with task-specific state variables and physical predicates.

The centrifuge bridge transforms calibrated socket mouths and directed insertion axes from a rotor-local frame to the world frame. It measures each tube's pose, velocity, release, and support. Insertion is confirmed jointly by radial clearance, axial alignment, depth, and continuous stability. The registered specification uses at most $10^\circ$ axial error, at least $60\%$ of effective socket depth, and a radial tolerance bounded by physical clearance. These geometry checks are conjoined with release and support over $0.5$\,s; final opposite occupancy persists for $3$\,s. Frame counts are derived from each configuration's physics rate.

The rack shares placement, release, and support measurements but uses a rectangular local slot map and a separate retrieval-choice recognizer. Its passive guides are confined to individual holes. Their controller handoff tolerances describe entry into a guide; the measured seating predicates determine successful placement. This separates placement assistance from the task's subsequent target-choice measurement.

\subsection{Force-derived numerical feedback}
For weighing, the evidence delivered to the policy must change with the load actually supported by the pan. The balance therefore uses an internal tray-support joint as its load cell and reads its measured wrench after physics stepping. The tray's sensing axis is checked against gravity, and an empty-pan calibration establishes the force bias before the unknown sample is introduced. With the support-force sign chosen opposite gravity, the conversion is
\begin{equation}
 m_t^{\rm raw}=\frac{1000}{g_0}
 \left(F_t^{\uparrow}-F_{\rm empty}^{\uparrow}\right),
 \label{eq:scale-readout}
\end{equation}
where force is in newtons, $g_0$ in $\mathrm{m/s^2}$, and mass in grams. A first-order filter drives the seven-segment readout:
\begin{equation}
 \bar m_t=\bar m_{t-1}+\alpha_t(m_t^{\rm raw}-\bar m_{t-1}),
 \quad \alpha_t=1-\exp(-\Delta t/\tau_f).
 \label{eq:scale-filter}
\end{equation}
The reference configuration uses $\tau_f=0.25$\,s, a $0.1$\,g display quantum, and a $1$\,s intermediate stability window with filtered range at most $0.5$\,g. Final completion combines a reading within $1$\,g of the goal, the required supported contents, release, low motion, and a continuous $3$\,s terminal window.

Mass configuration supplies evaluator-side calibration references; the live display is driven by the joint-force measurement. Contact and support measurements distinguish pan loading from gripper support or hand pressure. A finite force reading alone is insufficient to establish an eligible weighing state: the state update also checks release, support path, object motion, and overload status. Sensor errors invalidate the measurement window and are recorded with their source. The close display and weight-label cameras are additional named channels in the weighing profile.

\subsection{Generation, history, and environment transitions}
Resolved instances retain separate task-state, scene-layout, color or motion where applicable, and policy seeds. Feedback weighing samples an unknown preload and standard-weight subset, while the layout generator places the scale and labeled weights using a separate seed. The four denominations $10$, $20$, $50$, and $100$\,g have distinct subset sums. A stable reading and visible supported standards therefore identify the remaining feasible combination. Generation plans balance target and answer strata and stay on the evaluator side of the input boundary.

Placement recall records the actual initial slot and delivers its observations during the policy's own interaction. The \code{online\_once} history condition then delivers fresh current images after occlusion; the policy's declared session or memory component retains any earlier evidence. Same-cap distractors share the target's appearance, while distinct-cap distractors provide additional appearance cues. Color, local slot, and rack movement are balanced separately. These conditions measure retrieval under different evidence, with first slot choice and physical recovery scored independently.

The current hard rack condition uses an explicit occluded environment transition. After stable placement, safe retraction, and camera occlusion, it saves each tube's measured transform relative to the rack. Dynamic tube instances are reconstructed after the empty rack reaches its new transform:
\begin{equation}
 T'_{\rm world,tube}=T'_{\rm world,rack}\,T_{\rm rack,tube}.
 \label{eq:rack-rebuild}
\end{equation}
The transition preserves logical identity, appearance, and measured local pose; it initializes the recreated bodies at zero velocity and waits for physical stability. PhysX and USD transforms are checked before new frames are revealed. The environment mode and object generation identifiers enter the run record. Policy actions and operation recognition pause during this interval, while continuous video records the transition. Physical-transport diagnostics use a separate environment condition.

\subsection{Policy adapters and persistent records}
The shared policy layer provides HTTP, OpenPI, RPent-style agent, and VLA Evaluation Harness adapters. The OpenPI wrapper checks mapped image channels, prompt limits, action conventions, and reset declarations, and can return request and transformed-input receipts. The agent bridge exposes authorized observation, reporting, commitment, geometry, VLA or motion calls, and physical waiting through one episode driver. The harness adapter wraps that driver as evaluation infrastructure. A service enters a comparison after capability preflight and a physical online handoff for its declared interface.

Events retain time, object identity, evidence version, and both committed and measured targets. \code{run.json} stores resolved task and policy profiles, asset and contract hashes, all seeds, observation and intervention settings, and budgets. \code{events.jsonl} and \code{observations.jsonl} retain the measurements and delivered evidence. \code{online\_scores.jsonl} stores live snapshots, and \code{scores.json} contains the final report. Videos document the registered camera streams. Final scores are persisted before simulator closure, with encoder and application shutdown statuses recorded separately.

The demonstration recorder pairs the pre-action observation with the applied command and subsequent observation, retaining camera timestamps, action units, control frequency, and episode boundaries. Evaluator identities and scores occupy a separate diagnostic group. Environment-controlled intervals, such as an occluded rack transition, carry a control mask. Collection manifests retain failed and recovered attempts; the training index states its acceptance rule and split assignment. The HDF5 record and its training-format export preserve these identities.

Replay audit reconstructs state, legal choices, physical confirmations, and terminal dwell from saved measurements using the online scorer. It also checks that opportunities, intervention outputs, and subsequent actions share the appropriate evidence version. The audit verifies the consistency of the recorded online results.

\subsection{Validation and comparison records}
Each run records task rules, physical thresholds, observation access, environment transitions, and scoring denominators. Comparisons check these fields before matching episodes. Validation proceeds through geometry and sensor calibration, policy-interface handoff, complete physical execution, and matched model evaluation. Protocol tests exercise duplicate opportunities, stale evidence, shallow or unreleased placement, interrupted stability, unsupported interventions, and incomplete pairs. The corresponding records identify the conditions and measurements represented by each result.

\section{Experiments}
\label{sec:experiments}
\subsection{Feedback weighing as an initial test}
We begin with the easy feedback-weighing task, which couples a numerical choice to a complete manipulation sequence. An unlabeled sample is already on the scale. Given a target mass $m^\star$ and the observed stable reading $m_t$, the robot selects one of four standards, $\{10,20,50,100\}$\,g, to supply the residual $m^\star-m_t$. It then grasps, transports, releases, and verifies the selected standard. This setting tests whether successful object manipulation serves the mass requirement of the current instance. The initial study asks where this connection fails during full rollouts: choosing the required mass, realizing the transfer, or obtaining actions quickly enough for practical use. Shared-checkpoint and Oracle comparisons form the subsequent diagnostic study.

The task contract uses a force-derived scale reading. Completion requires the appropriate released load, a reading within 1\,g of the target, and a continuous 3\,s stable terminal window. The two VLA evaluations randomize weight layout, sample mass, and target mass. Each model is trained for 30,000 steps using 180 demonstrations and evaluated for 50 episodes, with a reported limit of 3,600 evaluation steps per episode. The task generator samples easy targets from 160, 180, and 200\,g and supplies a single-standard solution.

Changing the sample load and target changes the required residual mass and hence the correct standard. These randomized episodes therefore test state-dependent selection as well as manipulation across layouts. Their aggregate selection rate measures how often behavior meets the requirement of the current instance. Matched state pairs provide a complementary test by holding the goal and layout fixed while changing the load that determines the required choice.

\subsection{Evaluated systems and deployment conditions}
\paragraph{Fine-tuned VLA policies.}
We evaluate task-specific $\pi_{0.5}$ and GR00T-N1.5 policies. We report correct-weight selection, task completion, and mean runtime. Correct-weight selection is the fraction of evaluated episodes reported to select the standard required by the current residual mass. The available episode summaries provide this aggregate measure. Timestamped first choices, recognition coverage, and corrections belong to the event-level analysis defined in Section~\ref{sec:online}; the common-checkpoint metric in Eq.~\eqref{eq:bnext} additionally fixes the starting states across systems.

\paragraph{Additional system comparisons.}
The investigator-reported GPT-driven evaluation uses camera observations and robot state after removal of additional simulator geometry, and tests ten trials with varied initialization. The required standard is selected in all ten trials, and eight reach task completion. Table~\ref{tab:weighing-results} presents these system-specific evaluation populations alongside the two 50-episode VLA studies. Harness-VLA remains the pending agent-system comparison. Each system's delivered observations, action interface, controller, and computation budget define its evaluation condition. Matched-instance comparisons will fix these conditions within each base--support pair.

\begin{table}[t]
\centering
\caption{Initial feedback-weighing results. Selection and completion are separate episode-level measurements. The VLA studies randomize layout, sample mass, and target; the revised GPT study varies initialization. Counts specify each evaluation population. Harness-VLA is pending.}
\label{tab:weighing-results}
\fontsize{8.5}{10.5}\selectfont
\setlength{\tabcolsep}{3pt}
\renewcommand{\arraystretch}{1.12}
\begin{tabularx}{\columnwidth}{@{}Yrrr@{}}
\toprule
System & Trials & \shortstack{Correct\\selection} & Completion \\
\midrule
\multicolumn{4}{@{}l}{\textit{Randomized episodes; fine-tuned VLA policies}} \\
$\pi_{0.5}$ & 50 & 20\% & 16\% (8/50) \\
GR00T-N1.5 & 50 & 18\% & 8\% (4/50) \\
\midrule
GPT-driven control & 10 & 100\% & 80\% (8/10) \\
Harness-VLA & -- & -- & Pending \\
\bottomrule
\end{tabularx}
\end{table}

\begin{figure*}[t]
\centering
\begin{minipage}{0.485\textwidth}
\centering
\includegraphics[width=\linewidth]{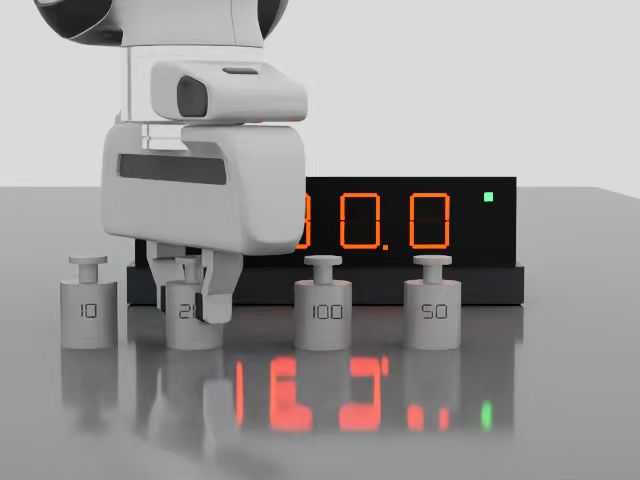}
\small (a) Upright grasp in a successful example
\end{minipage}\hfill
\begin{minipage}{0.485\textwidth}
\centering
\includegraphics[width=\linewidth]{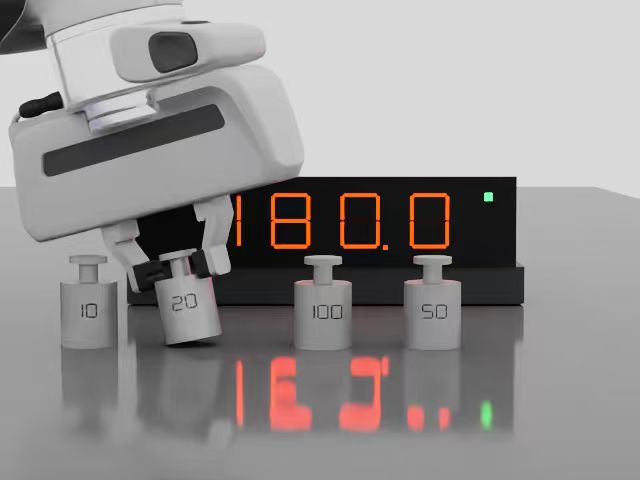}
\small (b) Tilted grasp in a failed example
\end{minipage}
\caption{Grasp configurations in the revised GPT evaluation. Both views concern the 20\,g standard. The right view shows a tilted object and gripper configuration. The evaluation descriptions associate failures with drops and report subsequent regrasp attempts, motivating separate measurement of target choice, grasp execution, and recovery.}
\label{fig:grasp-cases}
\end{figure*}

\subsection{Choosing a weight and executing its transfer}
\begin{figure*}[!tp]
\centering
\begingroup
\setlength{\tabcolsep}{3pt}
\newcommand{\vf}[2]{\begin{minipage}{0.316\textwidth}\centering\includegraphics[width=\linewidth]{media/weighing_#1.png}\\[-2pt]\footnotesize #2\end{minipage}}
\begin{tabular}{@{}ccc@{}}
\multicolumn{3}{@{}l}{\small\textbf{(a) $\pi_{0.5}$ success: 150\,g initial, 160\,g target; adds 10\,g}}\\[2pt]
\vf{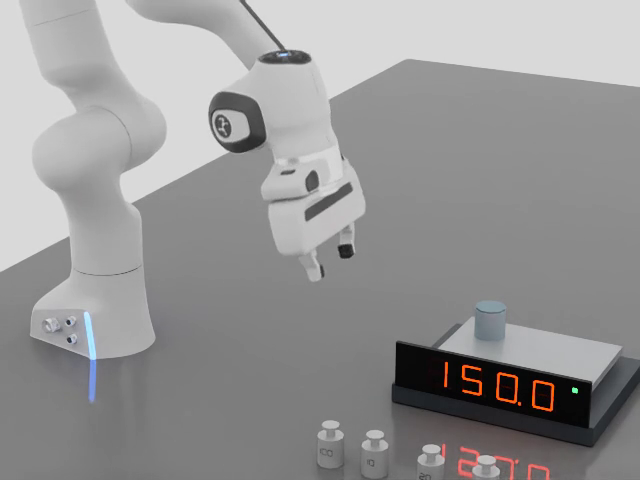}{3.20 s} & \vf{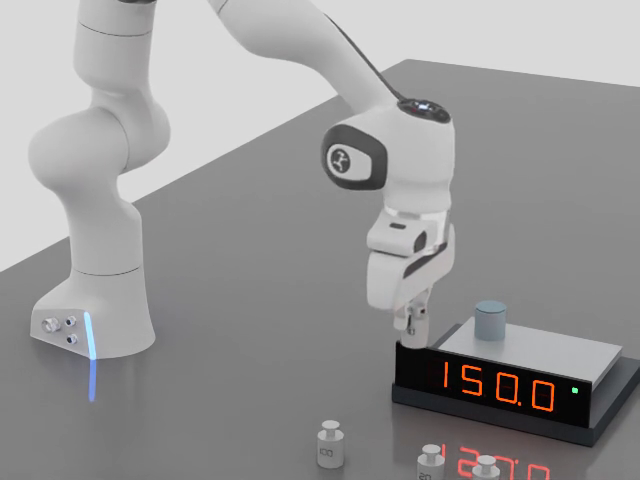}{12.70 s} & \vf{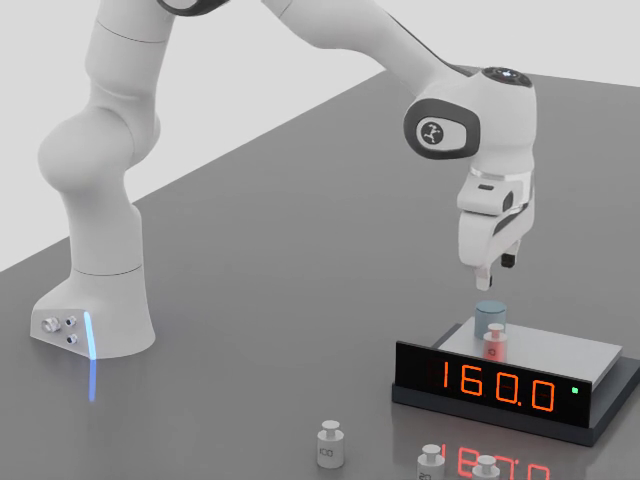}{20.80 s}\\[6pt]
\multicolumn{3}{@{}l}{\small\textbf{(b) $\pi_{0.5}$ failure: 170\,g initial, 180\,g target; adds 100\,g}}\\[2pt]
\vf{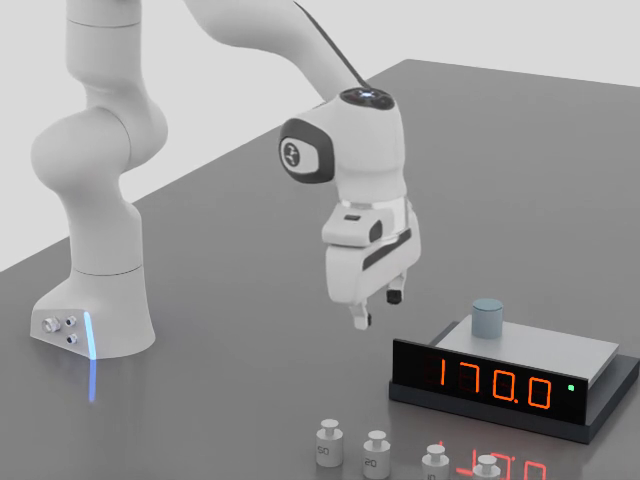}{4.50 s} & \vf{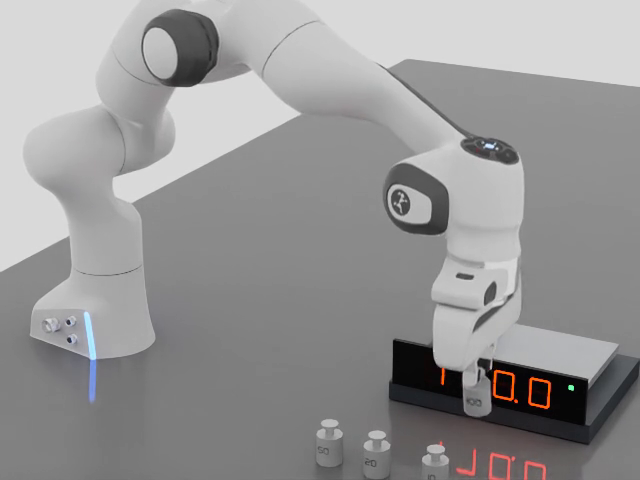}{13.50 s} & \vf{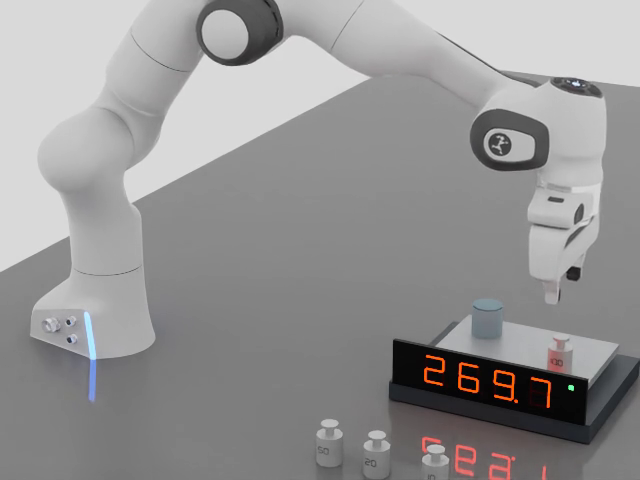}{29.30 s}\\[6pt]
\multicolumn{3}{@{}l}{\small\textbf{(c) GR00T-N1.5 success: 180\,g initial, 200\,g target; adds 20\,g}}\\[2pt]
\vf{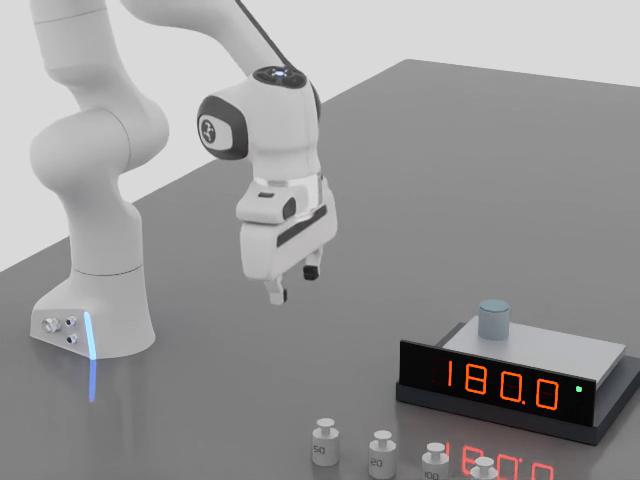}{5.00 s} & \vf{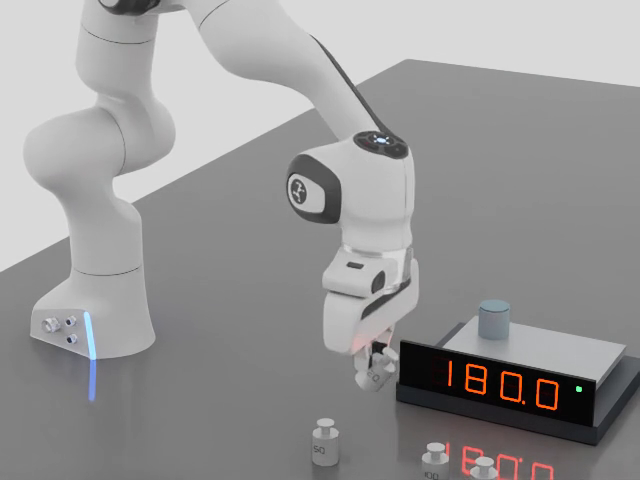}{20.00 s} & \vf{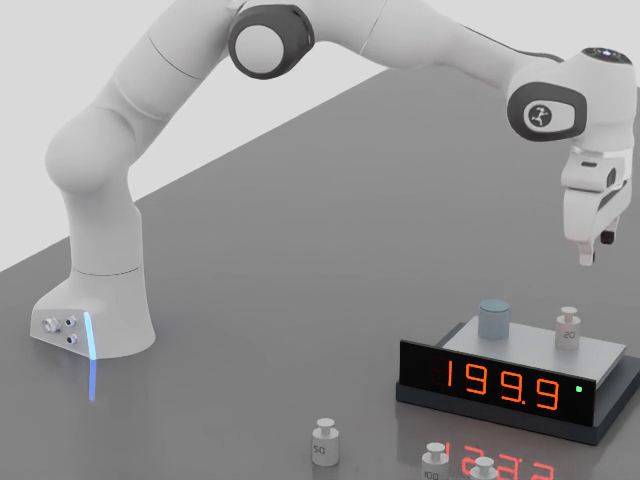}{32.60 s}\\[6pt]
\multicolumn{3}{@{}l}{\small\textbf{(d) GR00T-N1.5 failure: 170\,g initial, 180\,g target; adds 50\,g}}\\[2pt]
\vf{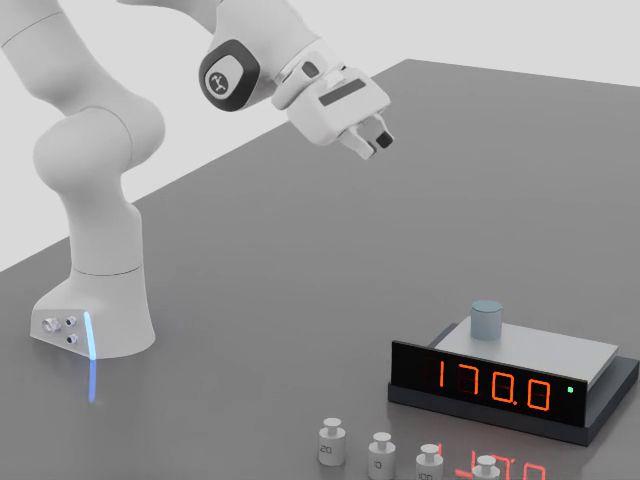}{1.20 s} & \vf{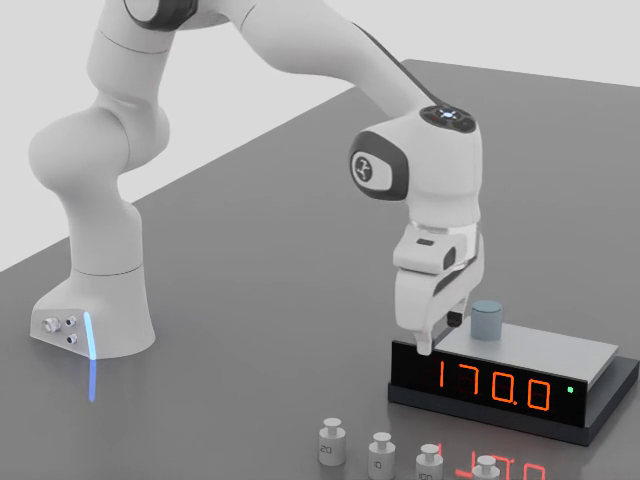}{18.00 s} & \vf{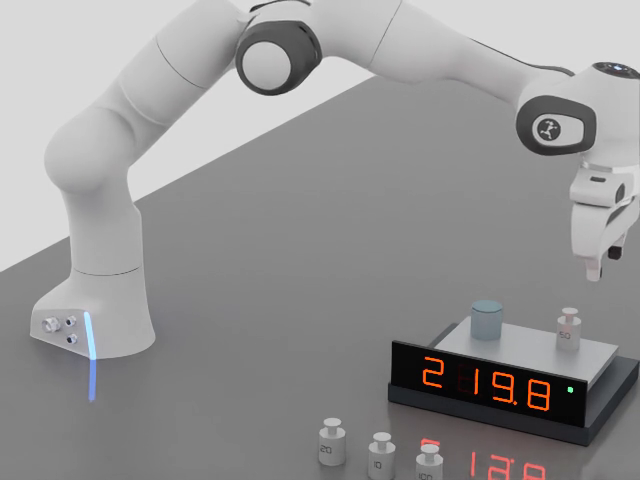}{58.70 s}
\end{tabular}
\endgroup
\caption{VLA weighing trajectories, read left to right within each row. Frames show the initial load, manipulation, and resulting load. Successful cases transfer the required standard; failed cases transfer a standard with the wrong mass. Targets and selected denominations follow the accompanying case descriptions. Timestamps are positions in the supplied videos, distinct from model-computation time. Images are extracted frames with their original field of view.}
\label{fig:vla-keyframes}
\end{figure*}

The VLA policies achieve correct-weight selection in 20\% and 18\% of episodes, while task completion reaches 16\% and 8\%, respectively. The distinction is visible in the recorded cases. In a successful $\pi_{0.5}$ episode, the initial reading is 150\,g and the target is 160\,g; the policy adds the 10\,g standard and completes the transfer. In a successful GR00T-N1.5 episode, it adds 20\,g to an initial 180\,g load, reaching a stable reading of 199.9\,g for a 200\,g target.

In the failed cases, successful transfer serves an inappropriate mass choice. For an initial reading of 170\,g and a target of 180\,g, $\pi_{0.5}$ transfers the 100\,g standard, while GR00T-N1.5 transfers the 50\,g standard. Although grasping and placement occur in both cases, neither selected mass satisfies the required 10\,g increment. The failure therefore concerns how the task requirement selects the operation, even when the chosen transfer can be executed.

Figure~\ref{fig:vla-keyframes} shows these four trajectories through initial-load, manipulation, and outcome frames. In the failed cases, the resulting readouts remain far above the requested 180\,g even after transfer. Appendix~\ref{app:physical-examples} separately presents keyframes from the physical-robot pilot recordings.

Qualitative observations across differing initializations also report repeated selection of the same standard as the required mass changes. This pattern indicates a fixed-weight preference in the observed runs: grasp-and-transfer behavior remains executable while object selection fails to track the numerical requirement. The preference is a behavioral observation; its frequency and dependence on denomination or spatial position will be resolved by the per-episode selection records.

Together, the low selection rates and recorded transfers show weak coupling between numerical task evidence and action selection in the tested fine-tuned policies. Training has produced executable manipulation behavior in individual episodes, while selecting the appropriate object remains a separate bottleneck. A VLM-based action policy must therefore be evaluated for whether its actions satisfy the current task state, alongside its ability to grasp and place objects. This distinction identifies the reasoning--action connection that \system{} is designed to measure.

\subsection{From a correct choice to a stable grasp}
Where the VLA examples expose incorrect object selection, the GPT trials illustrate the difficulty that can remain after the object is selected correctly. The required standard is selected in all ten trials, but eight reach task completion. The evaluation descriptions associate failures with grasp configurations that allow the weight to slip or fall, followed by regrasp attempts. An appropriate choice therefore shifts the evaluation question to its physical realization and recovery. Recovery attempts and completed recoveries will be counted separately from the corresponding trajectories.

Figure~\ref{fig:grasp-cases} shows successful- and failed-grasp examples involving the 20\,g standard. The failed example shows a tilted object and oblique gripper configuration, while the successful example shows an upright object between the fingers. The images illustrate the execution conditions underlying the reported cases. The aggregate result and these observations motivate a specific next test: given the same correct weight choice, which grasp and motion interface realizes that choice reliably?

\subsection{Runtime and deployment efficiency}
Successful transfer establishes physical progress; deployment additionally requires that progress at a useful timescale. The evaluation summaries report mean runtimes of 135.60\,s for $\pi_{0.5}$ and 86.81\,s for GR00T-N1.5. These aggregate episode-runtime records include the consequences of task outcome: a short episode may end in failure. Runtime is therefore interpreted alongside correct selection and completion.

The GPT evaluation reports about 30 minutes of model computation per trial after subtracting network delay, and about 50 seconds of task execution. These investigator-reported means place the dominant recorded time cost in computation. Observed pauses during target selection motivate finer timing of individual calls and operation stages. Correct selection, stable execution, and response time jointly determine how effectively a laboratory robot can advance through a procedure.

\paragraph{Capability and operational availability.}
The GPT system achieves correct weight selection and successful transfers in this setting. Its computation time makes timely access to that capability a central deployment requirement. For repetitive weighing, that cost directly affects throughput; for procedures involving evolving device states or time-sensitive handling, the response must also arrive while the observed conditions remain actionable. A reasoning--control interface must preserve both the intended operation and its timing requirements.

Interpreting this latency also requires the simulator's timing convention. In the synchronous runtime described in Section~\ref{sec:online}, physics pauses during inference, so the environment can wait for the next command. A continuously evolving physical process instead imposes a time window in which that command remains appropriate. Continued evaluation will therefore measure wall-clock time to completion and response latency after new evidence or failure, together with success under explicit response deadlines. These measurements test whether the demonstrated capability can be delivered quickly enough to sustain the procedure.

Timing records will separate model computation, communication, and physical execution, and include model-call and action-chunk counts. Stage-level timing will distinguish target selection, motion generation, and outcome verification. The reported aggregate computation time is a per-trial total; individual control-response latencies will be measured at their corresponding observation and action boundaries.

\subsection{Interpreting the reasoning--action connection}
The two evaluation patterns identify different points at which task progress can break down. The VLA failure examples realize transfers whose selected mass is inappropriate for the observed load. The GPT trials select the appropriate mass but sometimes fail to maintain the grasp needed to transfer it. A useful interface must carry the chosen object and operation into localization, grasp-pose generation, motion, and verification. After a drop, the target identity remains relevant while its pose and the next executable action must be updated. Regrasping therefore tests whether the same task requirement can be maintained through a changed physical state.

Taken together, these observations make the diagnostic question concrete: which support improves choice, physical realization, or timely continuation? The VLA cases motivate support for state interpretation and operation selection, whereas the GPT cases motivate support for grasp execution and efficient action generation. Matched checkpoints will test the corresponding transitions from common physical starts; execution-specific comparisons will additionally fix the correct selected subgoal.

\subsection{Limitations and continued evaluation}
The current empirical evidence consists of aggregate outcomes from randomized easy-weighing episodes and qualitative execution cases. This establishes a state-dependent selection test across initializations. Controlled pairs will isolate the effect of a specified state change under matched goals and layouts, using the joint-choice metric in Eq.~\eqref{eq:pair}. Event-level records will quantify first choices, corrections, identification coverage, and whether repeated selections follow a weight's denomination or position. Per-call timing will locate computation costs within selection, motion generation, and outcome verification.

The next comparisons will test which support improves this connection. For the VLA policies, state and decision support will examine whether the observed selection errors respond to a supplied readout or required symbolic addition. For GPT-driven control, common checkpoints with the same correct selected standard will compare candidate-geometry support and a validated execution controller, measuring grasp, transfer, release, and recovery. Each condition will be paired with the same system's base continuation, tracking choice changes, physical completion, and computation cost.

Harness-VLA will test an agent-mediated connection between high-level selection and the VLA execution policy. Its inputs, tools, and budgets will be specified alongside the native-policy and GPT configurations. Evaluation will extend to hard feedback weighing, centrifuge loading, and placement recall. Targeted improvements selected from development-set diagnostics will then be tested on held-out instances, with emphasis on residual-mass selection, outcome verification, and reduced model-call latency.

These extensions test whether a reasoning result remains an effective constraint on generated actions: the chosen mass must determine which weight is transferred, the chosen local slot must determine the insertion destination, and a remembered sample--slot association must determine retrieval after rack motion. Weighing provides the initial evidence for this connection; the spatial and history-dependent families provide the next evaluation settings.

\section{Conclusion}
\label{sec:conclusion}
Autonomous laboratories need robots that carry experimental requirements through physical execution. Selecting an appropriate operation is one step in that process: the resulting actions must preserve the required object, destination, quantity, and timing, and subsequent choices must reflect what the operation actually achieved. \system{} makes this reasoning--action connection the focus of evaluation. Its task families expose spatial, numerical, and history-dependent choices, with separate measurements of selection, execution, and completion. Shared checkpoints and matched Oracle continuations provide a protocol for testing which information or execution support helps restore progress.

Initial feedback-weighing results illustrate two different breakdowns. Across randomized initial states, the tested VLA policies select the required weight in 20\% and 18\% of their 50-episode evaluations, and complete 16\% and 8\%. Recorded failures include successful transfers of an inappropriate weight, and qualitative observations report repeated weight choices under changing requirements. These behaviors expose a gap between an executable manipulation skill and selecting its target from the current task state. The GPT-driven system selects correctly in all ten tested trials and completes eight; reported drops and regrasp attempts place the remaining difficulty downstream of a correct object choice. Its approximately 30-minute computation time per trial, compared with roughly 50 seconds of execution, also makes response efficiency central to practical use. Together, these observations motivate interfaces that preserve task requirements through action generation while delivering feedback quickly enough to sustain a procedure.

Matched Oracle studies will test how state information, operation decisions, geometry, and execution support change these outcomes. Extending the evaluation to centrifuge loading and placement recall will examine the same connection for specified insertion destinations and sample identity after relocation. These studies link diagnostic measurements to concrete improvements in how reasoning guides action.

This operational connection is a foundation for \emph{Embodied AI for Science}. Reducing hazardous, repetitive human work and sustaining autonomous exploration require experimental plans to become reliable physical procedures that produce interpretable scientific evidence. By measuring whether observed outcomes lead to appropriate choices and successful execution, \system{} provides a basis for improving that connection, from experimental intent to action and from realized results to the next operation.

\bibliographystyle{unsrtnat}
\bibliography{references}

\clearpage
\onecolumn
\appendix
\section{Laboratory Task Catalog}
\label{app:original-catalog}

The catalog contains twelve laboratory workflows (L01--L12) and twelve reasoning-in-action tasks (R01--R12). The following descriptions identify each task's operation, relevant evidence, and intended outcome. Simulator frames illustrate representative equipment and interactions. The three task families in Section~\ref{sec:families} develop spatial loading, quantitative feedback, and placement history in detail.

\newcommand{\taskentry}[2]{\paragraph{#1}#2}
\newcommand{\taskscene}[2]{%
\begin{minipage}[t]{0.32\textwidth}
\centering\includegraphics[width=\linewidth]{media/task_atlas/#1.png}
\par\vspace{3pt}{\small\textbf{#1}\quad #2}
\end{minipage}}

\subsection{Sample handling and preparation}

\begin{figure}[!ht]
\centering
\taskscene{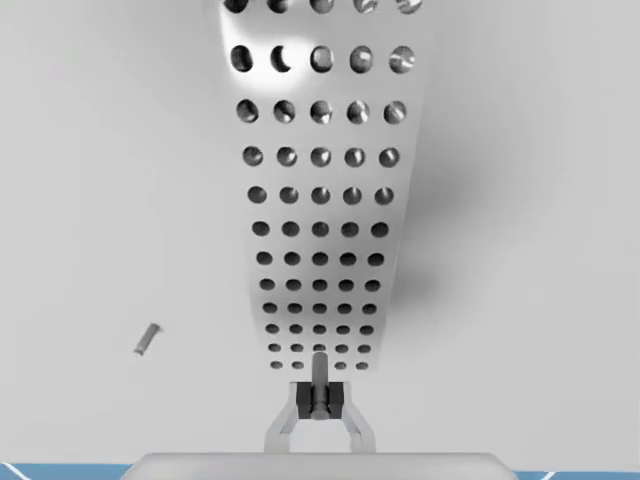}{Sample transport}\hfill
\taskscene{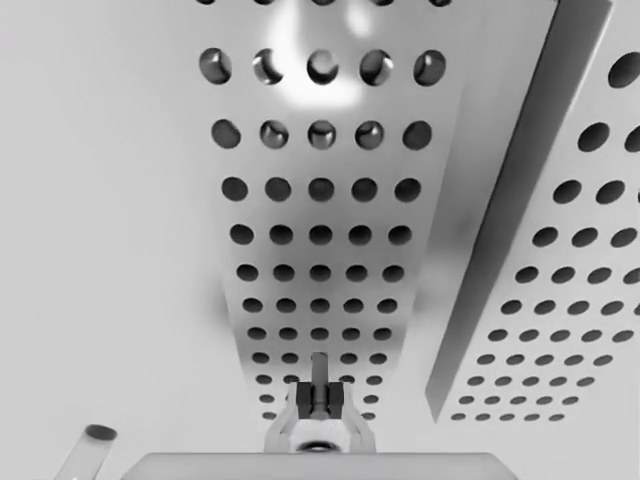}{Sample organization}\hfill
\taskscene{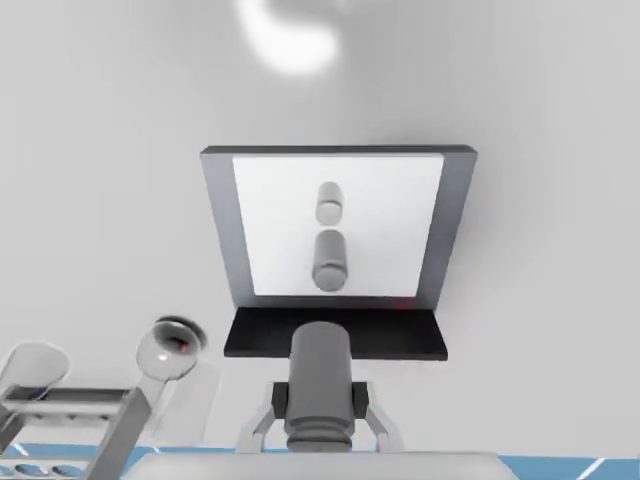}{Balance loading}
\caption{Simulator views of sample transport, sample organization, and balance loading. Wrist views expose the work area around the gripper.}
\label{fig:catalog-preparation}
\end{figure}

\taskentry{L01: Transport three sample vials.}{The robot picks three vials in sequence and places each in an assigned sample-box slot. It must retain the vial--destination association as the arrangement changes. The intended outcome is three released placements in their respective slots, with the requested order preserved.}

\taskentry{L02: Organize a vial by destination.}{Given a sample and an assigned destination, the robot identifies the corresponding box and moves the vial into its slot. Destination selection and released placement are distinct parts of the operation. This task supplies a sample-routing setting in which visually similar receptacles have different roles.}

\taskentry{L03: Tip installation and liquid transfer.}{The specified procedure installs a pipette tip, transfers $100\,\mu\mathrm{L}$ from a source tube to a destination, and discards the tip. The tool's tip state determines which operation can follow. The intended outcome combines the requested transfer, correct source and destination, and final tip disposal.}

\taskentry{L04: Three-beaker liquid mixing.}{Three source beakers are manipulated in turn, tilted toward a central receiving region, and returned to their starting places. Each source has its own pour-and-return obligation. The operation-level outcome records these three cycles and the final beaker arrangement.}

\taskentry{L05: Four-well recipe dispensing.}{A recipe assigns volumes to A1, A2, B1, and B2. The procedure follows the recipe while retaining which wells have already been processed. Correct well assignment, recipe coverage, and the absence of duplicate dispensing define the intended outcome.}

\taskentry{L06: Sequential balance loading.}{The robot places three labeled calibration weights on a balance in the specified order. It must distinguish the next weight from those already loaded and release each onto the pan. The outcome combines the loading sequence with the final supported weight arrangement.}

\clearpage
\subsection{Equipment use and workspace restoration}

\begin{figure}[!ht]
\centering
\taskscene{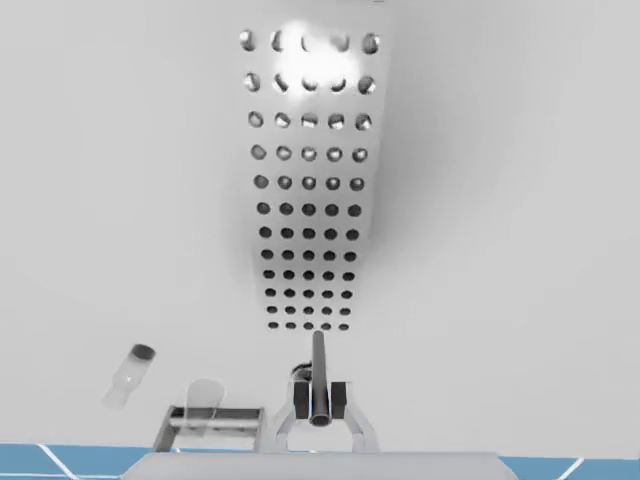}{NMR loading}\hfill
\taskscene{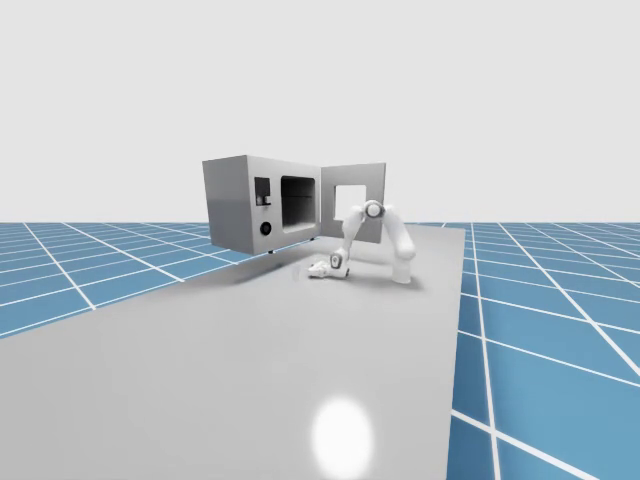}{Device loading}\hfill
\taskscene{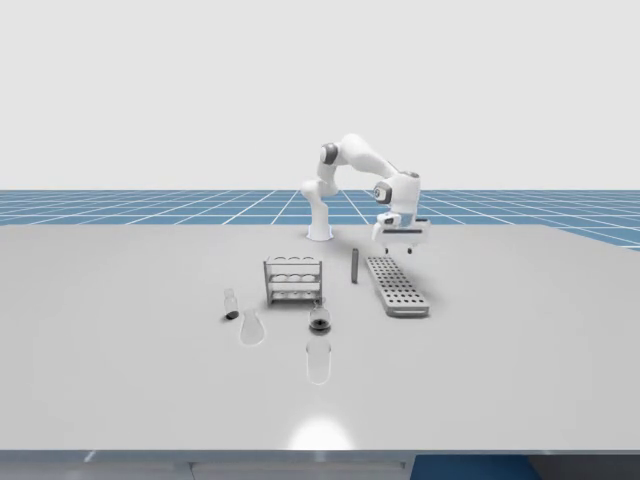}{Bench reset}
\caption{NMR sample handling, a device-loading interaction, and the workspace used for object return and reset.}
\label{fig:catalog-equipment}
\end{figure}

\taskentry{L07: Load two tubes into a centrifuge.}{Two tubes are picked and placed in designated rotor positions. The assignment of each tube and the occupancy established by the first placement guide the remaining loading action. The intended final arrangement contains both released tubes in the specified slots.}

\taskentry{L08: NMR sample-holder loading.}{The robot identifies the designated NMR tube, aligns it with its sample holder, and lowers it into place. Sample identity, target holder, and tube orientation jointly describe the goal. A seated and released tube completes the loading operation.}

\taskentry{L09: Load and start a laboratory device.}{The robot opens the device, places two beakers in its chamber, and actuates the physical start control. Access to the chamber and the loading state organize the sequence. The task connects object manipulation with an equipment-control action after loading.}

\taskentry{L10: Sequential three-vial loading.}{A queue assigns vials A, B, and C to positions 1, 2, and 3. The robot follows that queue while keeping track of completed placements. Correct identity--position associations and the specified ordering define the outcome.}

\taskentry{L11: Operate a condition module.}{The robot loads a beaker onto a condition module, activates its control, and returns the beaker after the cycle. The task connects handling before and after device use. Its intended sequence preserves the beaker's identity through loading, activation, and return.}

\taskentry{L12: Clear and reset the bench.}{A sample vial must return to its box and a test tube to its rack. The robot selects the corresponding destination for each object and clears the working area. Completion concerns the restored arrangement of all required objects, rather than the return of a single item.}

\clearpage
\subsection{Readouts, device states, and relational choices}

\begin{figure}[!ht]
\centering
\taskscene{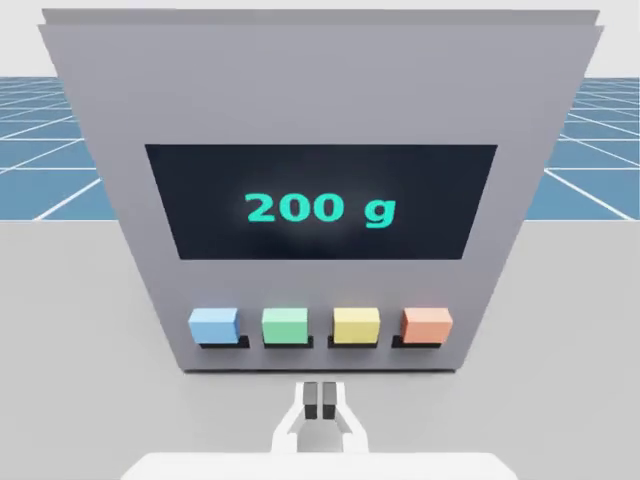}{Digital mass}\hfill
\taskscene{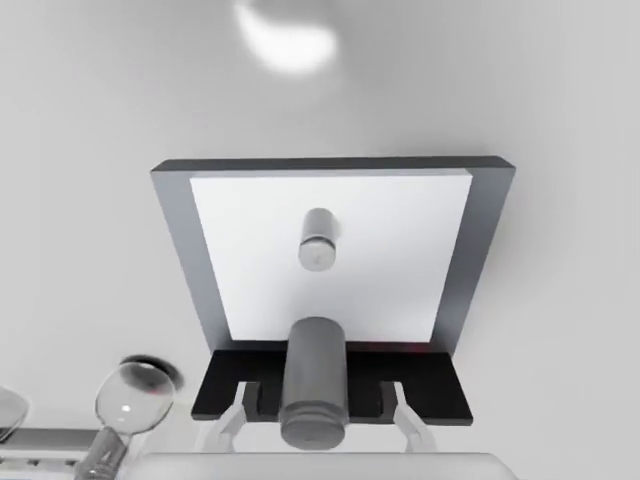}{Weight selection}\hfill
\taskscene{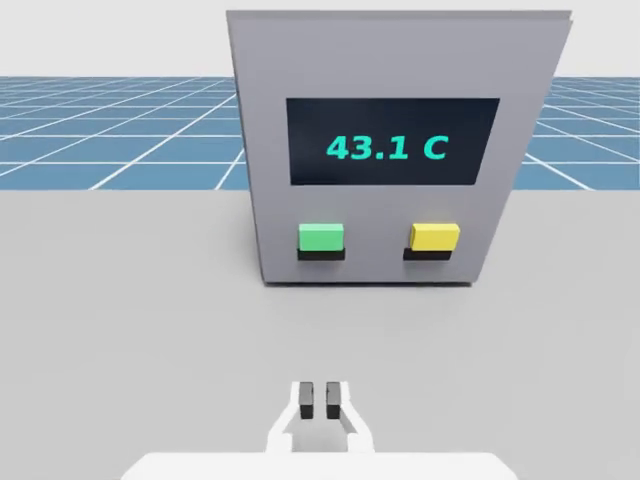}{Temperature panel}
\caption{Rendered evidence for numerical decisions: a mass display with candidate buttons, a weight-selection scene, and a temperature panel with action controls.}
\label{fig:catalog-readouts}
\end{figure}

\taskentry{R01: Read and record a digital mass.}{A scale display presents a mass and four buttons represent candidate values. The robot reads the number, selects its matching candidate, and presses that control. The task links a numerical observation to a discrete physical choice.}

\taskentry{R02: Select weights for a target mass.}{The robot reads standard-weight labels and chooses a combination for a stated target. It then places the selected standards on the balance. Combination correctness and the resulting supported load distinguish the selection problem from carrying out the placements.}

\taskentry{R03: Temperature-gated sample retrieval.}{A temperature display and the supplied $30^\circ\mathrm{C}$ threshold rule determine whether Wait or Open is appropriate. The robot applies the rule to the observed value and selects the corresponding control. Both the numerical interpretation and the chosen action affect the outcome.}

\taskentry{R04: RPM adjustment decision.}{The displayed speed is compared with a target RPM. The robot selects increase, decrease, or confirm according to the comparison. This is a directional adjustment decision: the observed difference determines which control should be actuated next.}

\taskentry{R05: State-dependent device control.}{The device presents Running or Done. Under the task rule, Running calls for Wait and Done calls for Open. The robot binds a visible process state to its corresponding control, so identical equipment can require different actions at different times.}

\taskentry{R06: Choose the opposite rotor slot.}{One rotor slot is occupied and another tube must be placed opposite it. The robot identifies the occupied location, derives its diametrically opposite partner, and selects that destination. Opposite-slot choice and the subsequent placement provide separate outcomes.}

\clearpage
\subsection{Occupancy, local coordinates, and access conditions}

\begin{figure}[!ht]
\centering
\taskscene{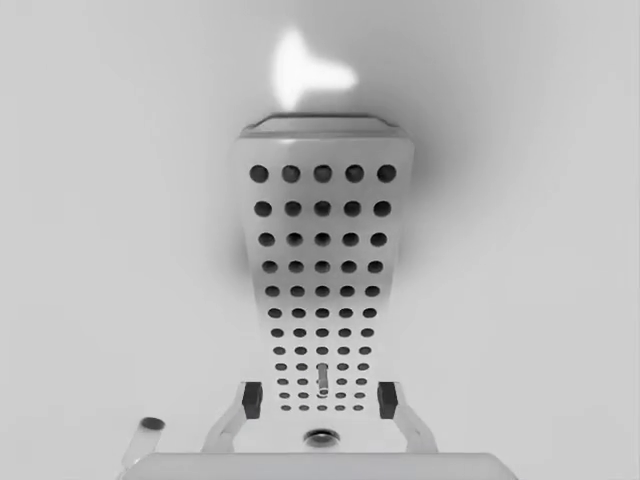}{NMR slot approach}\hfill
\taskscene{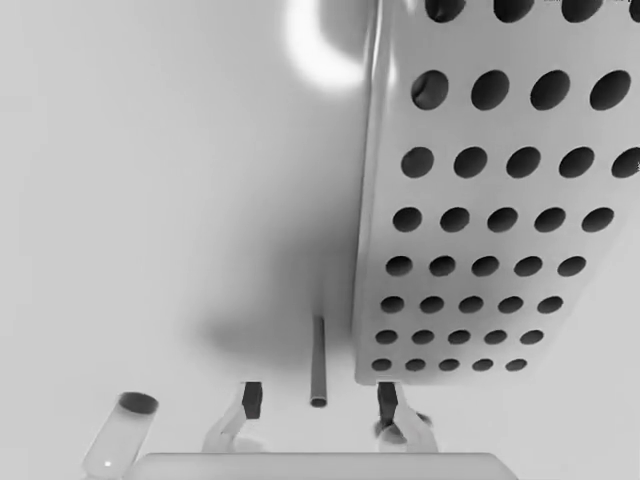}{Grid localization}\hfill
\taskscene{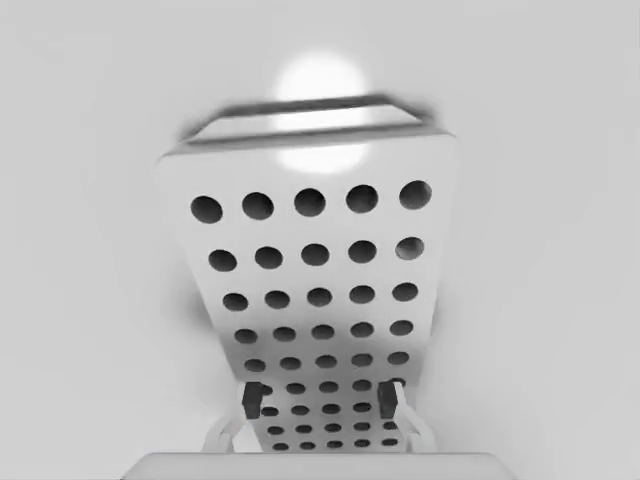}{Socket localization}
\caption{Wrist-camera views from slot-localization tasks. The images show the physical socket arrays approached by the gripper.}
\label{fig:catalog-slots}
\end{figure}

\taskentry{R07: Find an empty symmetric slot pair.}{The rotor is partly occupied. The robot must choose two available slots that are opposite one another and load a tube into each. The choice jointly depends on geometric symmetry and occupancy; both destinations must be available before loading begins.}

\taskentry{R08: Read and select an NMR slot.}{A display specifies a numbered destination, such as Slot 7. The robot connects that identifier to the corresponding physical socket and inserts the tube. The task binds a readout to a location in an array of similar destinations.}

\taskentry{R09: Map a B3 grid instruction.}{The instruction names a row and column on an NMR rack. The robot establishes the local grid directions, locates row B and column 3, and selects their intersection. The intended outcome is selection and placement at the commanded grid coordinate.}

\taskentry{R10: Place a tube at C2.}{A test-tube rack uses lettered rows and numbered columns. The robot maps C2 to its local socket and places the tube there. Correct local-coordinate interpretation and stable placement together define the goal.}

\taskentry{R11: Locate a sample-box socket.}{A row-and-column command specifies a socket in the sample holder. The robot identifies that local coordinate and moves the gripper above the corresponding opening. The outcome concerns the selected location and the gripper's arrival above it.}

\taskentry{R12: Lid-state action selection.}{The robot observes whether a device lid is open or closed. An open lid calls for Close; a closed lid permits Start under the task rule. The visible access condition therefore determines which operation is enabled next.}

\clearpage
\section{Physical-Robot Pilot Examples}
\label{app:physical-examples}

\begin{figure}[h]
\centering
\begingroup
\setlength{\tabcolsep}{5pt}
\newcommand{\rf}[2]{\begin{minipage}{0.245\textwidth}\centering\includegraphics[width=\linewidth]{media/weighing_#1.png}\\[-2pt]\small #2\end{minipage}}
\begin{tabular}{@{}ccc@{}}
\multicolumn{3}{@{}l}{\textbf{(a) Successful pilot recording}}\\[4pt]
\rf{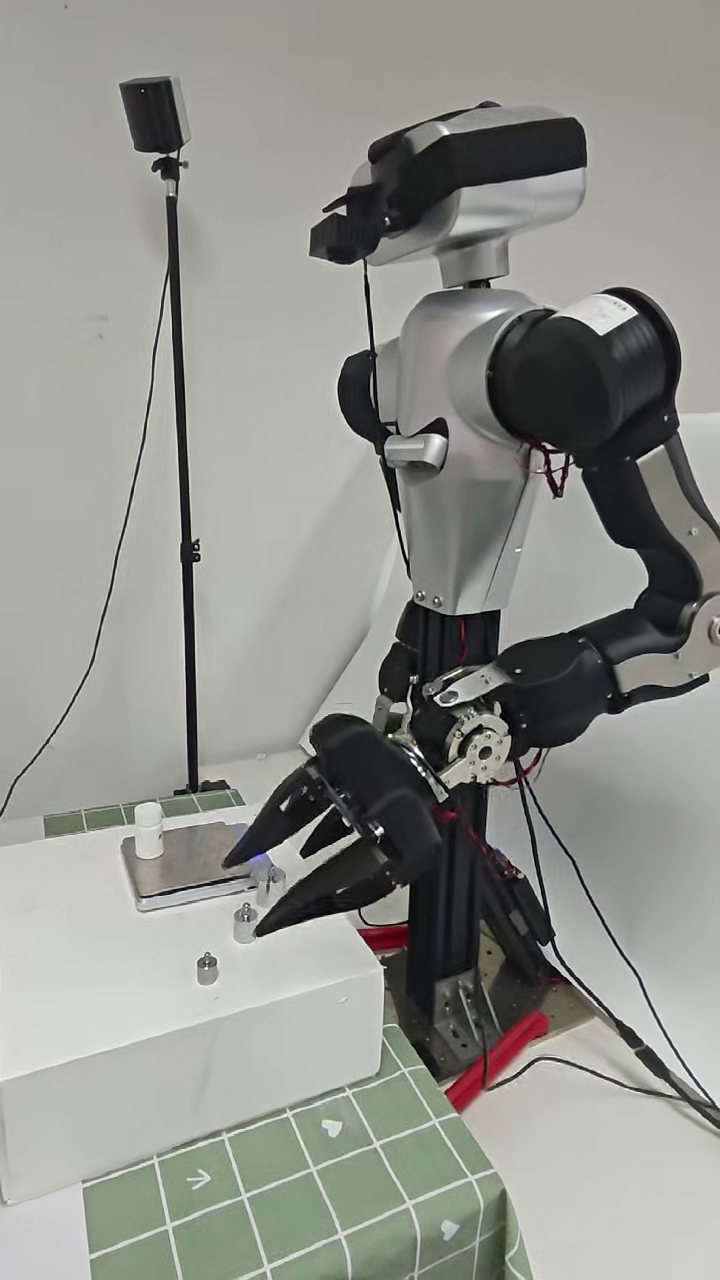}{1.83 s} & \rf{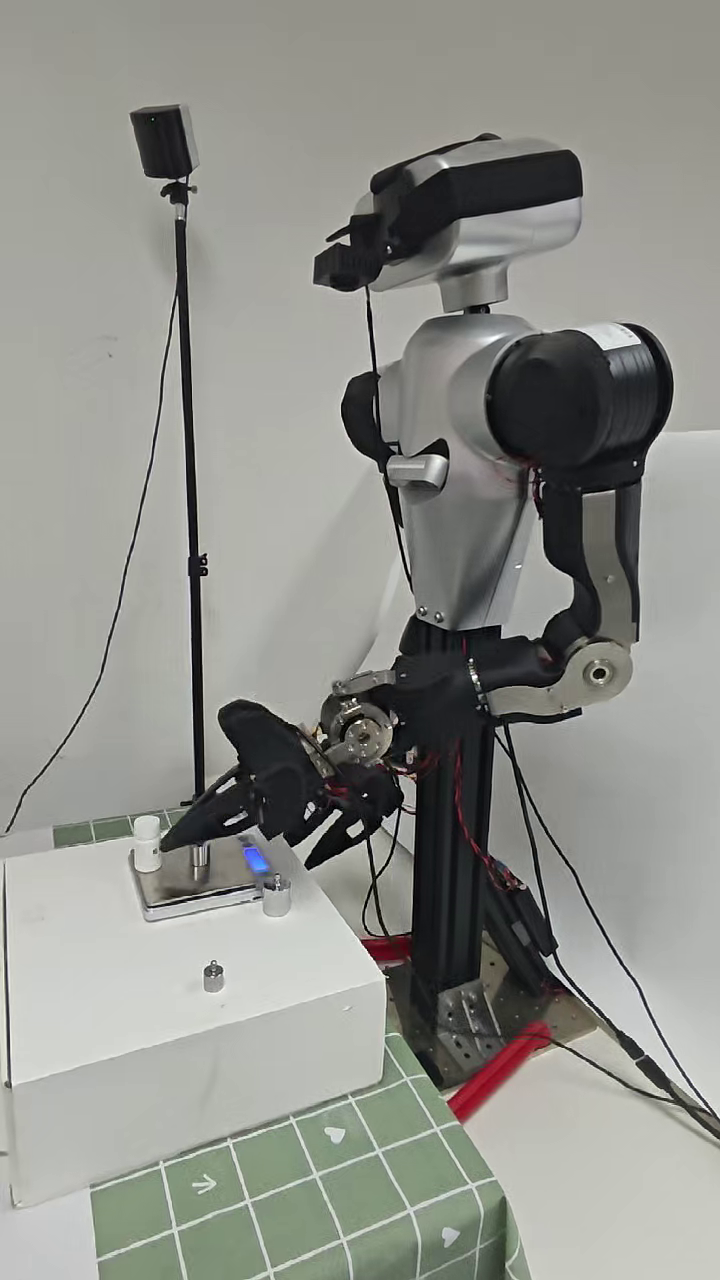}{7.33 s} & \rf{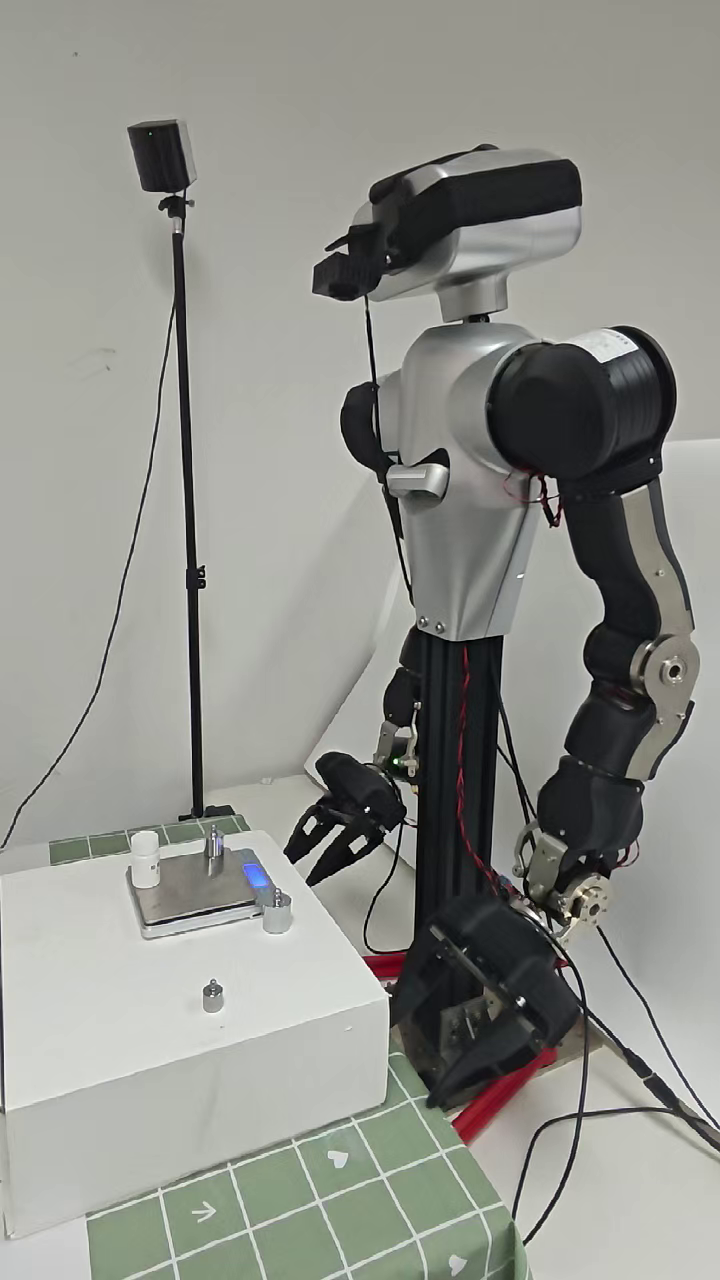}{12.00 s}\\[10pt]
\multicolumn{3}{@{}l}{\textbf{(b) Failed pilot recording}}\\[4pt]
\rf{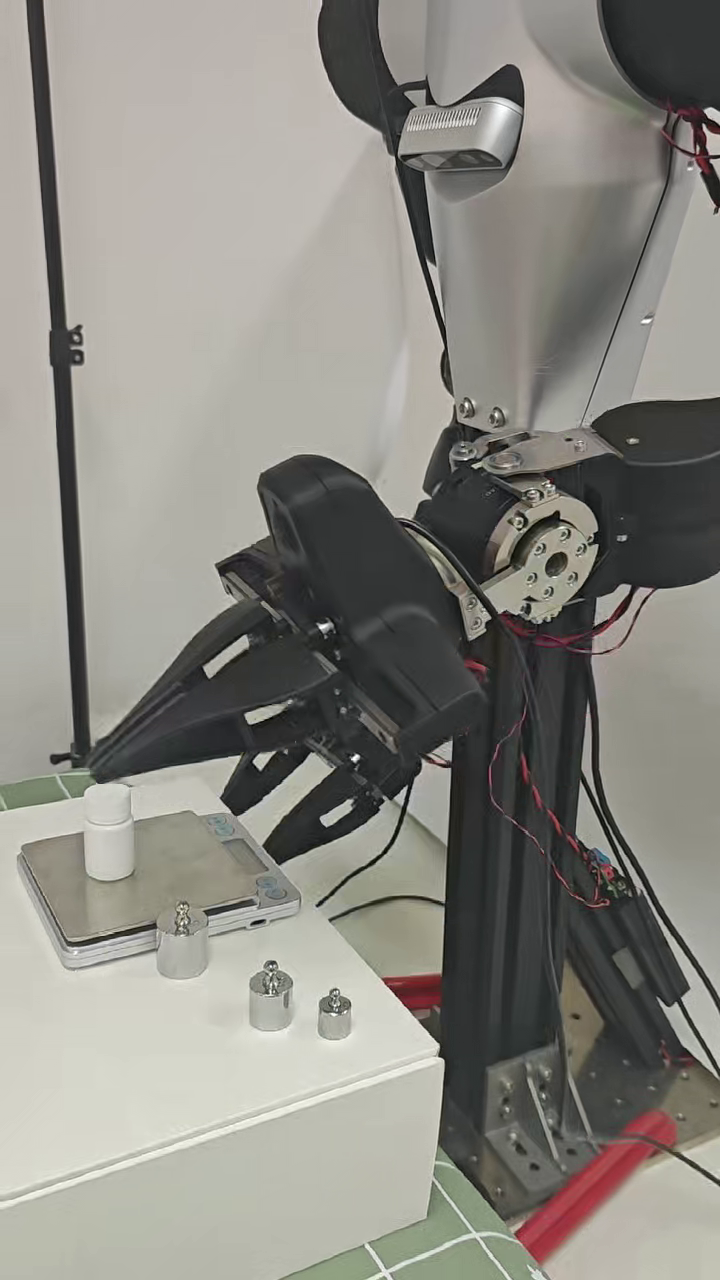}{5.60 s} & \rf{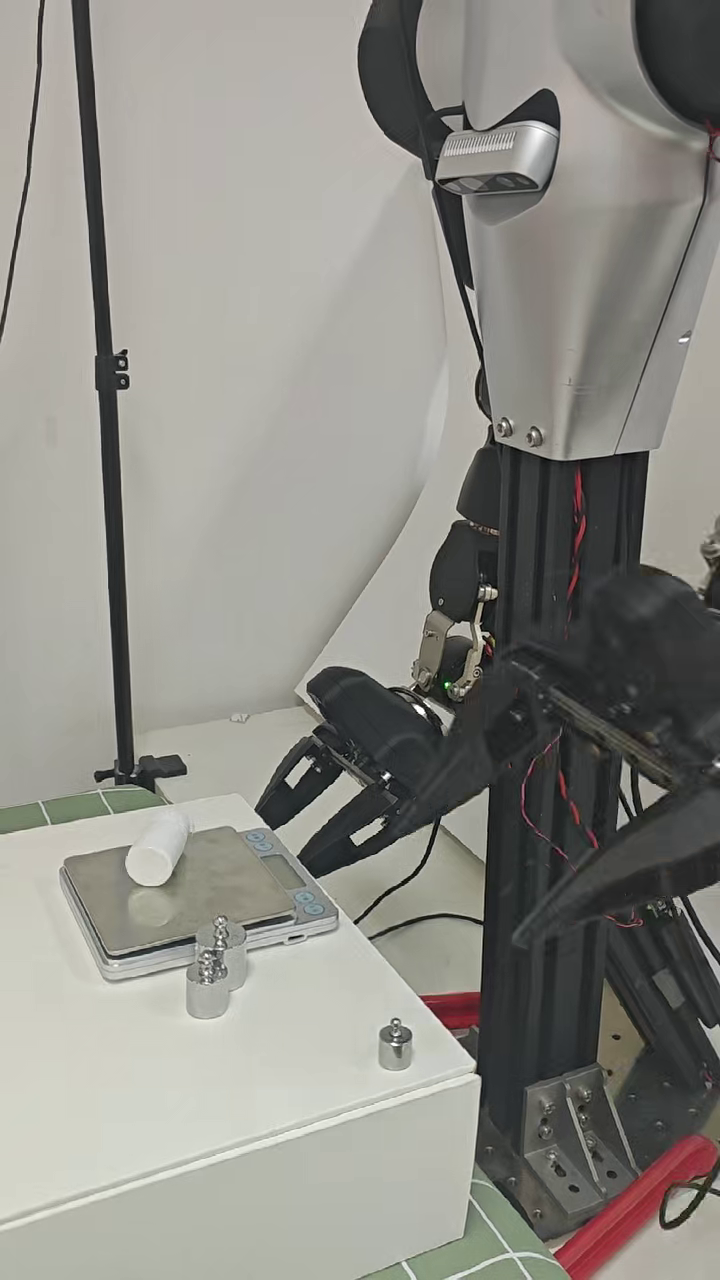}{22.47 s} & \rf{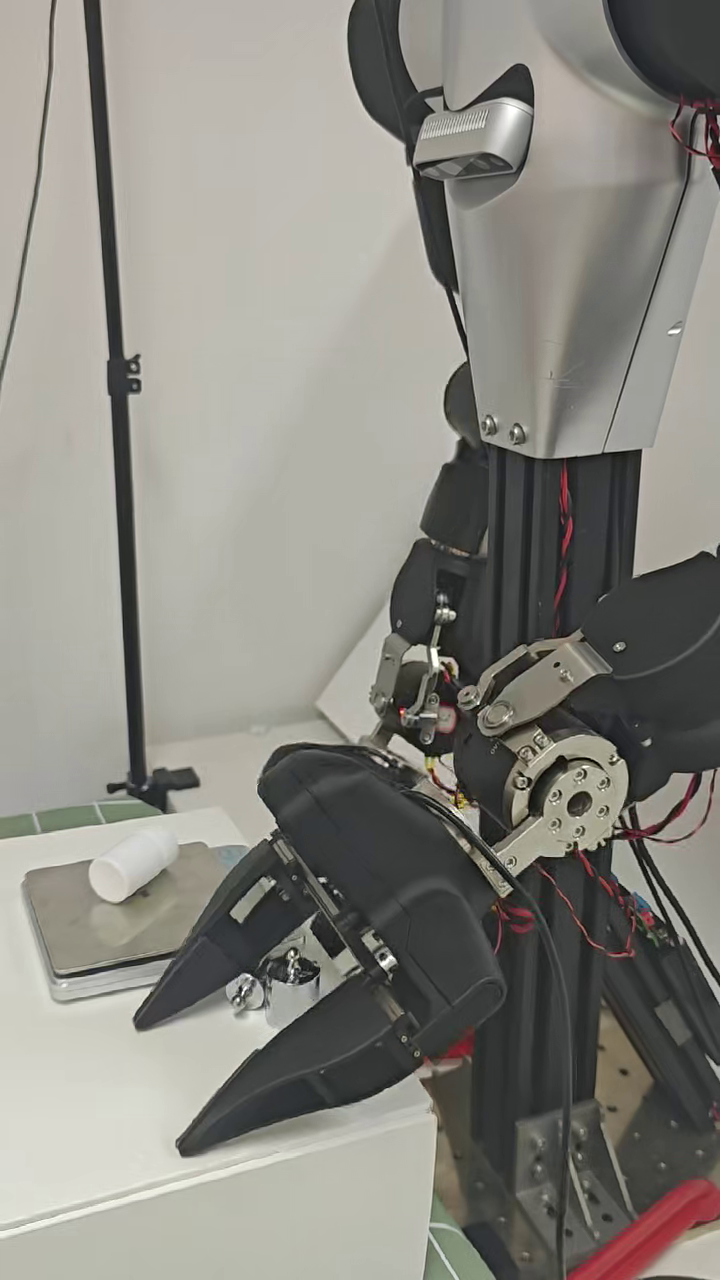}{36.70 s}
\end{tabular}
\endgroup
\caption{Physical-robot weighing pilot. Each row follows one recording in temporal order. Timestamps denote video positions. Original views are retained; the two recordings use different framing.}
\label{fig:physical-keyframes}
\end{figure}

Figure~\ref{fig:physical-keyframes} presents keyframes from the supplied successful and failed physical-robot recordings. These pilot examples document the real manipulation setup and observed interaction. Their outcome labels follow the experimental records. The quantitative results in Table~\ref{tab:weighing-results} concern the separate simulated easy-weighing evaluations.

\clearpage
\section*{Authors}
\addcontentsline{toc}{section}{Authors}

Shitong Liu\textsuperscript{1,2,\textdagger}, Xu Yang\textsuperscript{1,2},
Jingwei Zhang\textsuperscript{1,2},
Sheng Lian\textsuperscript{1,2}, Xinyu Zhuang\textsuperscript{1,2},
Kaiwen Guo\textsuperscript{1}, Quan Liu\textsuperscript{1},
Chunhai Zhao\textsuperscript{1}, Jiaxin He\textsuperscript{1},
Fan Yang\textsuperscript{1}, Manli Tao\textsuperscript{1},
Tingyu Yuan\textsuperscript{1},
Hongwei Shi\textsuperscript{3},
Chaomin Guo\textsuperscript{4}, Xiao-jing Liu\textsuperscript{5},
Feichan Yang\textsuperscript{6},
Zhuang Xiong\textsuperscript{7},
Xiangyu Zhu\textsuperscript{1}, Jinqiao Wang\textsuperscript{1,8},
Chaoyang Zhao\textsuperscript{1,9,\textdagger}, and Zhiyong Liu\textsuperscript{1,2,\textdagger}.

\medskip
\noindent\textsuperscript{1}Chinese Academy of Sciences Institute of Automation\\[3pt]
\textsuperscript{2}University of Chinese Academy of Sciences\\[3pt]
\textsuperscript{3}School of Pharmaceutical Sciences, Peking University\\[3pt]
\textsuperscript{4}Chinese PLA General Hospital\\[3pt]
\textsuperscript{5}School of Stomatology, Peking University\\[3pt]
\textsuperscript{6}IQuotient Robotics Co. Ltd.\\[3pt]
\textsuperscript{7}Institute of Semiconductors, Chinese Academy of Sciences\\[3pt]
\textsuperscript{8}Zidong Taichu (Beijing) Technology Co., Ltd.\\[3pt]
\textsuperscript{9}ObjectEye (Beijing) Technology Co., Ltd.

\end{document}